\documentclass[11pt]{article}

\usepackage[]{ACL2023}
\usepackage{booktabs}

\usepackage{multirow}
\usepackage{siunitx}
\usepackage{CJKutf8}
\usepackage{times}
\usepackage{latexsym}
\usepackage{microtype}
\usepackage{natbib}
\usepackage{booktabs}
\usepackage{appendix}
\usepackage{tikz}
\usepackage{tikz-dependency}
\usepackage{amsmath}
\usepackage{amsfonts}
\usepackage{graphicx}
\usepackage{amsthm}
\usepackage{mathtools}
\usepackage{multirow}
\usepackage{url}
\usepackage{color}
\usepackage{subfig}
\usepackage{dsfont}
\usepackage[ruled,vlined,noend]{algorithm2e}

\usepackage[noabbrev,capitalize]{cleveref}

\crefname{equation}{equation}{equations}
\crefname{footnote}{footnote}{footnotes}
\crefname{line}{line}{lines}
\crefname{section}{\S}{\S\S}
\Crefname{section}{\S}{\S\S}
\crefformat{section}{#2\S#1#3}
\Crefformat{section}{#2\S#1#3}
\crefrangeformat{section}{\S\S#3#1#4--#5#2#6}
\Crefrangeformat{section}{\S\S#3#1#4--#5#2#6}

\newcommand{\exper}[1]{\textsc{#1}}
\newcommand{\pamateur}{p_\exper{ama}}
\newcommand{\PMI}{\exper{PMI}}
\newcommand{\repn}{\exper{rep-n}}
\newcommand{\diversity}{\exper{div}}
\newcommand{\coherence}{\exper{coh}}
\newcommand{\mauve}{MAUVE}
\newcommand{\maxprob}{max prob}
\newcommand{\simctg}{CS\cite{Su2022ACF} }

\newcommand{\plm}{p_\exper{lm}}
\newcommand{\pexpert}{p_\exper{exp}}
\newcommand{\vv}{\mathcal{V}_\text{head}}

\newcommand{\Emb}{\exper{Emb}}

\newcommand{\xcont}{\textsf{x$_{\text{cont}}$}}
\newcommand{\xprompt}{\textsf{x$_{\text{pre}}$}}

\newcommand{\II}{I}

\DeclareMathOperator{\score}{CD-score}

\newcommand{\eg}{e.g.\ }
\newcommand{\ie}{i.e.\ }

\usepackage[T1]{fontenc}

\usepackage[utf8]{inputenc}
\usepackage[russian,british]{babel}

\usepackage{microtype}
\usepackage{inconsolata}

\title{Contrastive Decoding: Open-ended Text Generation as Optimization}

\author{Author 1 \and ... \and Author n \\
        Address line \\ ... \\ Address line}

\author{Xiang Lisa Li$^{1}$,  Ari Holtzman$^2$,  Daniel Fried$^3$, Percy Liang$^1$,  Jason Eisner$^4$,\\
{\bf Tatsunori Hashimoto$^1$, Luke Zettlemoyer$^{2,5}$, Mike Lewis$^5$} \\
Stanford University$^1$, University of Washington$^2$, Carnegie Mellon University$^3$, \\ Johns Hopkins University$^4$, FAIR$^5$\\
\texttt{xlisali@stanford.edu}, \texttt{ahai@cs.washington.edu}, \texttt{dfried@cs.cmu.edu}, \\\texttt{pliang@stanford.edu}, \texttt{jason@cs.jhu.edu}, \texttt{thashim@stanford.edu}, \\\texttt{lsz@cs.washington.edu},  \texttt{mikelewis@meta.com}
}

\begin{document}
\maketitle
\begin{abstract}
Given a language model (LM),  maximum probability is a poor decoding objective for open-ended generation, because it produces short and repetitive text. 
On the other hand, sampling can often produce incoherent text that drifts from the original topics. 
We propose contrastive decoding (CD), a reliable decoding approach that optimizes a contrastive objective subject to a plausibility constraint. 
The contrastive objective returns the difference between the likelihood under a large LM (called the expert, \eg OPT-13B) and a small LM (called the amateur, \eg OPT-125M), and the constraint ensures that the outputs are plausible.  
CD is inspired by the fact that the failures of larger LMs (e.g., repetition, incoherence) are even more prevalent in smaller LMs, and that this difference signals which texts should be preferred.
CD requires zero additional training,  and produces higher quality text than decoding from the larger LM alone. It also works across model scales (OPT-13B and GPT2-1.5B) and significantly outperforms four strong decoding algorithms (e.g., nucleus, top-k) in automatic and human evaluations across wikipedia, news and story domains.\footnote{Code is available at \url{https://github.com/XiangLi1999/ContrastiveDecoding.git}} 
\looseness=-1
\end{abstract}

\section{Introduction}
Open-ended text generation aims to craft fluent and coherent textual continuations of given prompts, laying foundations for various downstream applications such as writing assistance and story generation \cite{brown-et-al-gpt3}. The canonical approaches often sample from large pre-trained language models \cite{Holtzman2020Nucleus,fan-etal-2018-hierarchical, radford2019language}, but the generated text is prone to incoherence and topic drift as unlucky sampling choices compound over long sequences \cite{Eikema-2020-MAP,maynez-etal-2020-faithfulness}.
On the other hand, searching for the most likely sequences often results in short, repetitive and tedious text \cite{Holtzman2020Nucleus}, indicating that maximizing probability is a wrong decoding objective. 

\begin{figure}
    \centering
    \includegraphics[width=0.5\textwidth, page=1]{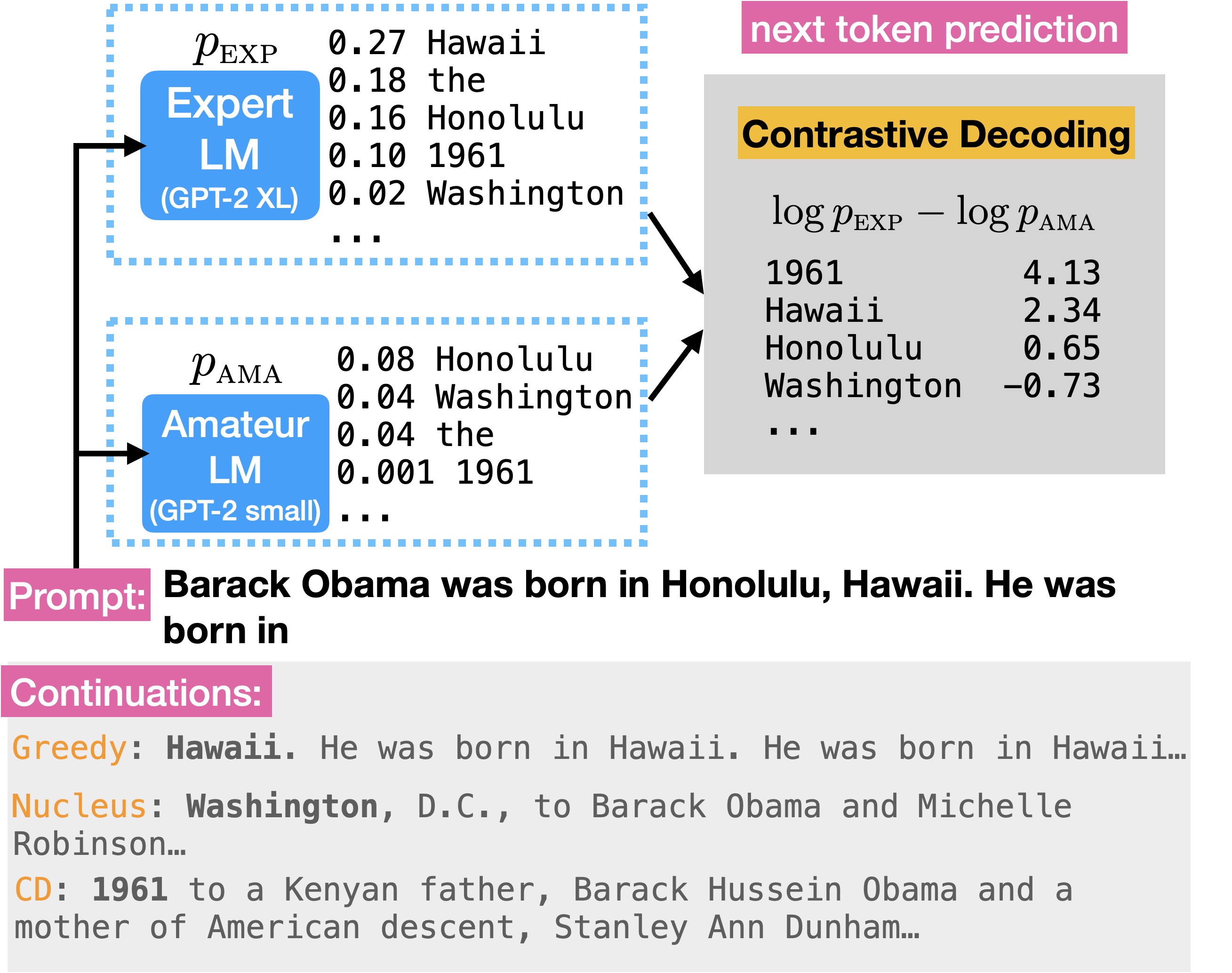}  
    \caption{\label{fig:fig1} Contrastive decoding exploits the contrasts between expert and amateur LM of different sizes by choosing tokens that maximize their log-likelihood difference. CD produces high-quality text that amplifies the good expert behavior and diminishes the undesired amateur behavior.
   \looseness=-1}
\end{figure}

We propose a new search-based approach, contrastive decoding (CD), that can generate fluent and lexically diverse 
text without compromising coherence. As shown in \cref{fig:fig1}, contrastive decoding takes an off-the-shelf large language model such as OPT-13B (that we call the expert) and an off-the-shelf smaller language model such as OPT-125M (that we call the amateur). CD \emph{searches} for text that \emph{maximizes} the difference between expert log-probabilities and amateur log-probabilities, subject to plausibility constraints which restrict the search space to tokens with sufficiently high probability under the expert LM. \looseness=-1

Contrastive Decoding works because many failure modes of language models (short, repetitive, irrelevant or uninteresting strings) are more common under smaller LMs than under larger LMs. Such outputs are further deemphasized by taking the difference between model log-probabilities. Conversely, stronger models tend to put more probability mass on desirable outputs, such as those with factual knowledge that has not been learnt by the weaker model, and these strings are emphasized by contrastive decoding.

Taking \cref{fig:fig1} as an example, the expert model places significant probability mass on previous tokens such as ``Hawaii'' and ``Honolulu'', leading to a highly repetitive continuation from greedy search; and nonsensical tokens such as ``Washington'' may be sampled, leading to an incoherent continuation. 
A correct continuation ``1961'' is strongly preferred by contrastive decoding, despite only having a probability of 0.1, and the continuation includes more correct facts.
This example suggests that contrastive decoding generates outputs that emphasize the best of the expert LM and remove its amateur tendencies. Moreover, we provide a pragmatic interpretation of contrastive decoding in \cref{sec:interpretation}. 

Compared to recent training-based methods that improve generation quality such as unlikelihood training \cite{Welleck2020Neural} and contrastive learning \cite{Su2022ACF, An2022CoNTCN}, contrastive decoding requires zero additional training. We find that by simply contrasting two \emph{frozen} language models of different sizes, we are able to decode higher quality text than from the larger LM alone. Furthermore, we find that better performance is achieved when the scale difference between expert and amateur is larger (\cref{sec:ablation_size}). As a result, the optimal amateur model is also cheap to run and incurs very little inference time overhead. 

We evaluate our contrastive decoding approach for open-ended text generation in three domains: Wikipedia, stories, and news, and we evaluate using different teacher-student combinations, including (GPT2-XL v.s. GPT2-small, OPT-13B v.s. OPT-125M). Compared to four decoding baselines (nucleus sampling, top-k, typical decoding and SimCTG) our contrastive decoding method significantly improves the coherence of generated text, and improves or maintains the same fluency levels, according to both human evaluation and automatic metrics.  

\section{Problem Statement}
We consider decoding approaches for open-ended language generation, where the language models receive an input prompt and aim to generate a fluent and coherent continuation. 
Specifically, we consider a relatively short prompt of length $n$, denoted as $\xprompt = x_1 \cdots x_n$, where $x_i$ is a token in the vocabulary $\mathcal{V}$. The decoder must generate continuations of length $m$, denoted as $\xcont = x_{n+1}, \cdots, x_{n+m}$. 

We generate text from a pre-trained autoregressive language model $\plm$.  At decoding time, we iteratively decode one token at a time by conditioning on the preceding context:%
$$ \plm (\xcont \mid \xprompt)
= \prod_{i=n+1}^{n+m} \plm (x_i \mid x_{<i}).$$ 
\noindent
where $\plm (x_i \mid x_{<i})$ is the next token distribution. We use different subscripts to denote different LMs: $\pamateur$ is the amateur LM (e.g., GPT-2 small), and $\pexpert$ is the expert LM (e.g., GPT-2 XL). 

One canonical decoding approach is to sample from a truncated next token distribution at each time step. For example, nucleus sampling \cite{Holtzman2020Nucleus}  draws from the top $p$ percentile of the next token distribution; top-k sampling \cite{fan-etal-2018-hierarchical} draws from the top $k$ candidates in the next token distribution. Another common approach is to search for the most likely text sequence via greedy decoding or beam search \cite{wu2016google}; but this leads to repetition and tedious outputs.

\section{Contrastive Decoding} 
We propose contrastive decoding as a search-based decoding method that optimizes a novel contrastive objective subject to our plausibility constraint. We first provide intuition and define the constrastive objective (\cref{ssec:intuition}). Second, we discuss the potential weakness of this objective alone, and introduce the plausibility constraint to correct for the weakness (\cref{ssec:plausibility}). Then we define the full contrastive decoding method as our contrastive objective subject to the plausibility constraint (\cref{ssec:contrastive}). Finally, we elaborate on the design spaces by discussing the choices of amateurs (\cref{ssec:amateur}). 

\subsection{Contrastive Objective}
\label{ssec:intuition}
Smaller LMs demonstrate stronger tendencies to produce 
undesirable patterns (e.g., repetition, topic drift, and self contradiction) than larger LMs. For example, when both expert (larger LM) and amateur (smaller LM) assign highest probability to a repetitive token, the expert LM is often less confident about this decision and assigns non-trivial probability mass to other good, non-repetitive continuations. 
Contrastive decoding is inspired by these observations. The goal is to factor out undesired behaviors highlighted by the smaller amateur LMs, and generate text from the remaining good behaviors of larger expert LMs. 

To operationalize this intuition, we propose the contrastive objective $\mathcal{L}_{\text{CD}} (\xcont, \xprompt)$:
\begin{align*}
    \log \pexpert (\xcont \mid \xprompt) - \log \pamateur (\xcont \mid \xprompt)
\end{align*}

The CD objective rewards text patterns favored by the large expert LMs and penalizes patterns favored by the small amateur LMs. However, amateur LMs are not always mistaken: small language models still capture many simple aspects of English grammar and common sense (e.g., subject verb agreement). Thus, penalizing all behaviors from amateur LMs indiscriminately would penalize these simple aspects that are correct (False negative), and conversely reward implausible tokens (False positive).  To tackle this issue, we introduce the plausibility constraint, which complements our CD objective and avoids these failure modes. 

\subsection{$\vv$: Adaptive Plausibility Constraint} 
\label{ssec:plausibility}

To tackle the aforementioned issue, we propose an adaptive plausibility constraint ($\vv$) that exploits the confidence level of the expert LM to restrict the effect of the contrastive objective when the expert LM is highly confident: 
\begin{align}
& \vv (x_{<i}) = \\ 
& \{x_i \in \mathcal{V}: \pexpert(x_i \mid x_{<i}) \geq \alpha \max_w \pexpert(w | x_{<i}) \} \nonumber
\end{align}
Here, $\alpha$ is a hyperparameter in $[0, 1]$ that truncates the next token distribution of $\pexpert$. Larger $\alpha$ entails more aggressive truncation,  keeping only high probability tokens, whereas smaller $\alpha$ allows tokens of lower probabilities to be generated. We set $\alpha=0.1$ throughout the paper.

This adaptive plausibility constraint corrects for both false positive and false negative failures of the contrastive objective: 
\paragraph{False positives.} An implausible token may be rewarded with a high score under our  unconstrained contrastive objective. For example, the token ``NetMessage'' is highly implausible under the context of \cref{fig:fig1}, with \num{3e-9} of $\pexpert$ and \num{8e-14} of $\pamateur$; however, it attains the highest contrast of $\log \pexpert - \log \pamateur = 10.6$, which is much higher than plausible tokens ``1961'' and ``Hawaii''. To handle the false positive problem, $\vv$ filters out low probability tokens and only keeps high probability tokens in the candidate pool.

\paragraph{False negatives.} When confronting an easy decision, the correct token that achieves high probability under both amateur LM and expert LM may receive a low score under the contrastive objective. For example, due to tokenization, the word ``unicorn'' consists of two subwords: ``unic'' and ``\#orn'', and the probability of ``\#orn'' given the prefix ``unic'' is close to 0.99 under both LMs, but the contrast $\log \pexpert - \log \pamateur$ is only \num{6e-4}, which is much lower than bad continuations. 

Here, $\vv$ uses the expert LM's confidence (as defined by the $\alpha$ ratio with the max probability token in the given timestep) to avoid these false negative cases. 
The expert LM assigns high confidence to easy decisions, but not to tokens that reflect the undesired behaviors of the amateur, since probability mass is taken up by other candidate tokens the expert is able to consider. 
Our constraint keeps as few as one token in the candidate pool when the expert is highly confident about this token, which removes the impact of the contrastive objective, because the single token would always be highest ranked regardless of the CD objective. 

\subsection{Full Method}
\label{ssec:contrastive}
Combining the contrastive objective and the adaptive plausibility constraint, we obtain the full contrastive decoding formulation: 
\begin{align}
    & \max_{\xcont} \mathcal{L}_{\text{CD}} (\xcont, \xprompt) \\
    & \text{subject to ~~~ }  x_i \in \vv(x_{<i}), \forall x_i \in \xcont \nonumber
\end{align}

The above objective is defined at the sequence level, which is intractable to optimize. Thus, we factor the objective to token level scores:
\begin{align}
& \score(x_i; x_{<i}) \label{eqn:cont}\\ 
& = \begin{cases}
\log \frac{\pexpert ( x_i \mid x_{<i})}{ \pamateur( x_i \mid x_{<i})},
& \text{if } x_i \in \vv(x_{<i}) \text{,}\\
-\inf,    	      & \text{otherwise.} 
\end{cases}  \nonumber 
\end{align}

We apply beam search to optimize $\score$, by first filtering tokens based on plausibility constraints $\vv(x_{<i})$, eliminating tokens that fail to achieve sufficiently high probabilities under the expert LM. Then we score the remaining tokens based on the amount of contrast they demonstrate, according to $ \log \pexpert ( x_i \mid x_{<i}) - \log \pamateur( x_i \mid x_{<i})$. As a result, we end up selecting plausible tokens under the expert LM that least resemble the amateur LM. 

\subsection{Choice of Amateur}
\label{ssec:amateur}
The choice of amateur LM is an important decision for contrastive decoding. 
As discussed in \cref{ssec:intuition}, we should choose amateur LMs that exhibit the behaviors we would like to downweight from the expert LM. Here, we consider three aspects:

\paragraph{Scale.}
Smaller LMs have lower modeling capacity and are more prone to errors. Therefore, we choose the amateur LM to be the smallest model in the same family of the expert LM. For example, for OPT-13B expert, we choose OPT-125M as the amateur; for GPT-2 XL expert, we choose GPT-2 small as the amateur. We verify this design choice in \cref{sec:ablation_size}. 
On the extreme end, employing n-gram models yields an amateur LM of extremely low capacity. But this choice hurts generation quality, because n-gram LMs incur too many errors to identify similar failure modes of the expert LM. 

\paragraph{Temperature.}
We can manipulate the amateur LM behavior by tuning its temperature $\tau$. For example, applying a high temperature ($\tau > 1$) to the amateur LM results in flatter distributions; applying a low temperature ($\tau$ close to $0$) highlights the mode of the amateur distribution, which is more prone to errors (\eg repetition). Therefore, we manipulate the temperature of the amateur LM to adjust the amateur behavior that will be penalized in contrastive decoding. In \cref{ssec:ablahyper}, we study the impact of $\tau$ to generation quality and set $\tau$ to $0.5$ or $1.0$ for our main experiments.  

\paragraph{Context window.}
We can also weaken capacity by restricting the context window of the amateur LM \cite{li-etal-2016-diversity}. For instance, we can only allow the amateur LM to condition on the last token of $\xprompt$, but we allow the expert LM to condition on the entire $\xprompt$. 
In other words, we decode from $\log \frac{\pexpert(\xcont \mid x_{1:n})}{\pamateur(\xcont \mid x_n)}$. By conditioning the amateur LM only on partial prompts, the coherence of the amateur LM is weakened, and contrastive decoding produces more coherent text by highlighting the coherence nature of the expert LM. In \cref{ssec:abla_prompt}, we study the impact of this design choice. 

\section{CD as Pragmatic Communication}
\label{sec:interpretation}
Having formally described contrastive decoding, we now provide a pragmatic interpretation, justifying its validity through pragmatic communication goals \looseness=-1. 

A line of work in pragmatics~\citep{grice1975logic} characterizes communication as a cooperative process between speakers and listeners. Several of these formalisms~\citep{Horn1984-implicature,Levinson2000-implicature} describe a tradeoff between speakers and listeners, where a speaker should generally produce language that is high quality (e.g. truthful, fluent, and relevant) while also being informative to a listener.

Our contrastive objective can be motivated by this tradeoff, with our expert and amateur LMs modeling a knowledgable speaker and a less-informed listener: 
(1) Upweighting tokens by $\pexpert$ and using our expert-based plausibility constraints generates tokens that have high probability under the expert LM, encouraging generated text to be fluent and relevant 
(\eg upweighting `1961' in \autoref{fig:fig1}). 
(2) Downweighting tokens by $\pamateur$ suppresses language that is predictable by (\ie less informative to) the amateur LM (e.g. downweighting `Honolulu' and `Washington'), and by proxy encourages the language to be informative to a listener in context.
By combining these two criteria, our contrastive decoding method produces high quality text that satisfies the communicative goal of transferring relevant but not predictable information. 

\subsection{Special Cases of Contrastive Decoding} 

\paragraph{Maximum probability.} Setting the amateur LM to a uniform distribution reduces CD to maximize log-probabilities under the expert LM. 

\paragraph{N-gram blocking.} If we set the amateur LM as an n-gram model whose n-gram counts are updated to fit the generated prefix, this yields a decoding algorithm with soft n-gram blocking. If we also set the amateur temperature to be very small, then it approaches the canonical heuristic of forbidding repeated n-grams \cite{paulus-2018-a}. 

\paragraph{Diverse decoding.} If we use the same LM as both amateur and expert and restrict the context window of the amateur LM (\cref{ssec:amateur}), our method is equivalant to the MMI decoding objective \cite{li-etal-2016-diversity} sometimes used in dialog systems, which explicitly maximizes the pointwise mutual information between the $\xprompt$ and $\xcont$.

\section{Experimental Setup} 
\label{sec:setup}

\subsection{Datasets and Metrics} 
\label{ssec:metrics}
We evaluate on three domains for open-ended text generation: news, Wikipedia, and story domains. 
For the news domain, we use news articles from Wikinews;\footnote{Wikinews from \url{ http://www.wikinews.org}} for the Wikipedia domain, we use the WikiText-103 dataset \cite{merity2017pointer}; and for story domains, we use the BookCorpus \cite{moviebook} (Project Gutenberg split).  

We use the first 32 words in the passage as the prompt, and  decode for 256 tokens for the continuations.   
We evaluate generated text with both automatic and human evaluation. 

\paragraph{Diversity.} This metrics aggregate n-gram repetition rates: 
$\diversity = \prod_{n=2}^4 \frac{|\text{unique n-grams (\xcont)}|}{\text{total n-grams (\xcont)}|}$.
A low diversity score suggests the model suffers from repetition, and a high diversity score means the model generated text is lexically diverse. 

\paragraph{MAUVE.} MAUVE \cite{pillutla2021mauve} score (the higher the better) measures the distribution similarity between the set of generated text and the set of gold reference. 

\paragraph{Coherence.} We follow \citet{Su2022ACF} and approximate coherence by cosine similarity between the sentence embeddings of prompt $\xprompt$ and generated continuation $\xcont$: $\coherence (\xcont, \xprompt) = \frac{\Emb(\xprompt) \cdot \Emb(\xcont)}{||\Emb(\xprompt)|| \cdot ||\Emb(\xcont)||}$, where $\Emb(x)$ is the pre-trained SimCSE sentence embedding \cite{gao2021simcse}. \looseness=-1

\paragraph{Human Eval.}
In order to evaluate the quality of the generated text, we consider two critical aspects: \emph{fluency} and \emph{coherence}. A fluent piece of text is written in grammatical English and has a natural flow (\eg excluding unnatural repetition or web formatting). A coherent piece of text should stay on topic with the prompt and avoid unnatural topic drift.
We ask Amazon Mechanical Turkers to read two continuations (A and B) of the same prompt, and  choose the more fluent/coherent continuation or decide they are similar. 

\subsection{Baselines}
We compare contrastive decoding with three sampling methods, each with the recommended hyperparameters: nucleus sampling ($p=0.95$), top-k sampling ($k=50$), typical decoding \cite{Meister-2022-typica} ($\tau=0.95$); and two search-based methods: greedy (max prob) decoding that uses $\log \pexpert$ as the objective, 
and contrastive search (CS) \cite{Su2022ACF, su2022contrastiveiswhatyouneed}.
Among them, nucleus sampling is the standard approach for open-ended text generation whose performance has been verified in various domains \cite{Holtzman2020Nucleus,DeLucia-2020-decoding}, and typical decoding is a recently proposed approach that excels in lexical diversity \cite{Meister-2022-typica}. We therefore conduct human evaluation by comparing CD against these two methods. 

\subsection{Models and Hyperparameters} 
In order to demonstrate that our approach generalizes across various LM families and sizes, we consider GPT-2 XL (1.5B), OPT (6.7B) and OPT (13B) as expert LMs and employ the smallest LM in their respective family as the amateurs: GPT-2 small (100M) and OPT (125M). 

Recall that contrastive decoding introduces two hyperparameters: 
$\alpha$ is the parameter to adjust the plausibility threshold, and $\tau$ is the temperature of the amateur LM. We always set $\alpha=0.1$ for the main results in the paper --- we find that this setting is quite robust and generalizes across various domains. For OPT experiments, we set the amateur temperature to $1.0$ and for GPT-2 experiments, we set the amateur temperature to $0.5$. We use a beam size of 5. We also study the impact of these hyperparameters in the ablation study \cref{ssec:ablahyper}, and we find that our method is robust to various hyperparameter values. \looseness=-1

\section{Main Results} 

\subsection{Automatic Evaluation}
As shown in \cref{tab:automatic_eval}, contrastive decoding outperforms all other decoding baselines in MAUVE score and coherence score (\coherence) across three different domains (news, Wikipedia, stories) and two model sizes (1.5B, 13B). Contrastive decoding achieves comparable or slightly worse diversity compared to nucleus and typical sampling, but it achieves substantially better diversity than other search based methods. 

Typical decoding and nucleus sampling produce lexically diverse text by choosing low probability tokens, at the expense of topic drift. For instance, in the story domain we observe the largest diversity gap between contrastive decoding and nucleus sampling (0.83 v.s. 0.94) in the 1.5B model, but we find that the gap shrinks (0.89 v.s. 0.93) as the model size increases to 13 billion, suggesting that our decoding method would continue to improve as expert models continue to scale. 

CD outperforms all the baselines in coherence scores by a large margin, followed by greedy decoding. Greedy decoding achieves good coherence despite being highly repetitive, because always repeating the same sentence is a degenerate way to circumvent topic drift. We believe our gain in coherence comes from three aspects: (1) CD searches to optimize our objective, avoiding the topic drift that can happen by chance in sampling-based generation techniques.
(2) Our contrastive objective implicitly rewards coherence, because large LMs are typically more coherent than smaller LMs.
(3) Finally, we restrict the context length of the amateur LM (\cref{ssec:amateur}), further encouraging CD to reward text that is connected with the prompt \cite{li-etal-2016-diversity}. \looseness=-1

\begin{table*}[]
\centering
\resizebox{0.8\textwidth}{!}{
\begin{tabular}{llccc|ccc|cccllllll}
& & \multicolumn{3}{c}{wikinews} & \multicolumn{3}{c}{wikitext} & \multicolumn{3}{c}{story} \\
& name&     \diversity &    \mauve &   \coherence & \diversity & \mauve & \coherence  &  \diversity &    \mauve&    \coherence & \\
\toprule
\multirow{7}{*}{\rotatebox{90}{OPT-13B}} 
& \maxprob&                       0.08&   0.3&    0.65 &  0.03&   0.08&   0.63 & 0.02&   0.05&   0.51& \\
& k=50&                         0.91&   0.92&   0.64 &  0.72&   0.77&   0.64 & 0.91&   0.9&    0.51& \\
& p=0.95&                       0.92&   0.92&   0.62 & \textbf{0.92}&   0.89&   0.55  & 0.93&   0.91&   0.48& \\
& typical=0.95&                 \textbf{0.94}&   0.9&    0.59 & 0.89&   0.86&   0.58 & \textbf{0.95}&   0.91&   0.46& \\
& \simctg  &   0.92&   0.87&   0.59 &  0.87&   0.77&   0.52  & 0.81&   0.78&   0.47& \\
& CD &       \textbf{0.94}&   \textbf{0.94}&  \textbf{ 0.69} &  0.91&   \textbf{0.91}&   \textbf{0.69} &  0.89&  \textbf{0.94}&   \textbf{0.62}& \\
\midrule
\multirow{8}{*}{\rotatebox{90}{GPT2-XL}} 
& \maxprob &                      0.04&   0.14&   0.65 &  0.02&   0.05&   0.62 & 0.01&   0.03&   0.49& \\
& k=50&                        0.92&   0.88&   0.64  &  0.87&   0.79&   0.61  & 0.91&   0.87&   0.51& \\
& p=0.95&                   0.94&   0.9&    0.6&   0.92&   0.87&   0.57 & 0.94&   0.91&   0.46& \\
& typical=0.95&                \textbf{0.95}&   0.91&   0.56  &    \textbf{0.95}&  0.84&   0.53 & \textbf{0.96}&   0.88&   0.43& \\
& \simctg  &             0.93&   0.82&   0.62 &   0.86&   0.75&   0.59 & 0.88&   0.78&   0.48& \\
& CD &         0.92&   \textbf{0.94}&   \textbf{0.69}  & 0.89&   \textbf{0.92}&   \textbf{0.69}& 0.83&   \textbf{0.94}&  \textbf{0.64}& \\

\end{tabular}
}
\caption{\label{tab:automatic_eval}
Automatic evaluation results for wikipedia, wikinews, story datasets. The best scores for each (model, domain) setting are boldfaced. Contrastive decoding outperforms all other decoding baselines in MAUVE score and coherence score (\coherence) for different model scales (1.5B, 6.7B, 13B). CD achieves comparable or slightly worse diversity compared to nucleus and typical sampling. 
}
\end{table*}
\subsection{Human Evaluation}
We conduct human evaluation to compare our contrastive decoding approach against nucleus sampling (the canonical method that scores high under MAUVE) and typical decoding (the winning method for diversity metrics).\footnote{%
Prior work has found that these methods outperform other proposed decoding algorithms \cite{DeLucia-2020-decoding,Meister-2022-typica}}

As shown in \cref{tab:human_eval}, contrastive decoding generates significantly more coherent text compared to nucleus and typical decoding across three domains and two models: on average across settings, evaluators preferred CD 2.6x more than nucleus sampling and 6.4x more than typical decoding when evaluating coherence. 
As for fluency, CD is preferred 1.4x more than nucleus sampling and 3.5x more than typical decoding. 

\begin{table*}[]
\centering
\resizebox{1.0\textwidth}{!}{
\begin{tabular}{lcc|ccc|ccccccllllll}
& & &  \multicolumn{3}{c}{coherence} & \multicolumn{3}{c}{fluency} \\
& CD & Baseline  & CD is better &     same  &    Baseline is better &  CD is better &     same  &    Baseline is better & \\

\toprule

\multirow{4}{*}{\rotatebox{90}{wikitext}} 
& CD (GPT-2 XL) &  nucleus (GPT-2 XL) & \textbf{0.714}$^*$ & 0.083 & 0.202 & \textbf{0.548 } & 0.083 & 0.369 \\
& CD (GPT-2 XL) & typical (GPT-2 XL) & \textbf{0.887}$^*$ & 0.046 & 0.067 & \textbf{0.703}$^*$ & 0.082 & 0.215 \\ 
& CD (OPT-13B) & nucleus (OPT-13B) & \textbf{0.556 } & 0.202 & 0.242 & \textbf{0.419 } & 0.197 & 0.384 \\ 
& CD (OPT-13B) & typical (OPT-13B) & \textbf{0.773}$^*$ & 0.106 & 0.121 & \textbf{0.687}$^*$ & 0.152 &  0.162 \\
\midrule
\multirow{4}{*}{\rotatebox{90}{wikinews}} 
& CD (GPT-2 XL) &  nucleus (GPT-2 XL) & \textbf{0.708}$^*$ & 0.042 & 0.25 & \textbf{0.583}$^*$ & 0.12 & 0.297 \\ 
& CD (GPT-2 XL) & typical (GPT-2 XL) & \textbf{0.771}$^*$ & 0.151 & 0.078 & \textbf{0.755}$^*$ & 0.151 & 0.094 \\ 
& CD (OPT-13B) & nucleus (OPT-13B) & \textbf{0.585}$^*$ & 0.221 & 0.195 & \textbf{0.518 } & 0.123 & 0.359 \\ 
& CD (OPT-13B) & typical (OPT-13B) & \textbf{0.693}$^*$ & 0.099 & 0.208 & \textbf{0.49 }& 0.297 & 0.214 \\ 
\midrule
\multirow{4}{*}{\rotatebox{90}{story}} 
& CD (GPT-2 XL) &  nucleus (GPT-2 XL) & \textbf{0.636}$^*$ & 0.045 & 0.318 & 0.404 & 0.106 & \textbf{ 0.49} \\ 
& CD (GPT-2 XL) & typical (GPT-2 XL) & \textbf{0.506 } & 0.256 & 0.238 & \textbf{0.387} & 0.363 & 0.25 \\ 
& CD (OPT-13B) & nucleus (OPT-13B) & \textbf{0.616}$^*$ & 0.101 & 0.283 & \textbf{0.449} & 0.293 & 0.258 \\ 
& CD (OPT-13B) & typical (OPT-13B) & \textbf{0.626}$^*$ & 0.202 & 0.172 & \textbf{0.52} & 0.212 & 0.268 \\ 
\bottomrule

\end{tabular}}
\caption{\label{tab:human_eval} Human evaluation results for wikipedia, wikinews, story  datasets.  We boldfaced the max column and * the numbers that are significantly better than the two other columns combined. 
Contrastive decoding generates significantly more coherent text compared to nucleus and typical decoding across three domains and two models. CD also generates better or comparably fluent text compared to two baselines. }
\end{table*}

\subsection{Qualitative Examples} 

\begin{table*}
\small 
    \centering
    \resizebox{0.95\linewidth}{!}{
    \begin{tabular}{p{0.8cm}p{14cm}}
\toprule
Prompt &   A version of Sonic the Hedgehog was developed by Ancient and released in 1991 for Sega's 8 - bit consoles, the Master System and Game\\ 
\midrule
CD (1.5B) &   A version of Sonic the Hedgehog was developed by Ancient and released in 1991 for Sega's 8 - bit consoles, the Master System and Game Gear. It featured a new level editor called the Sonic Editor and was released as an expansion pack to Sonic CD in 1992, which was itself a port from the Genesis version. The editor allowed the creation and playtesting of custom levels for Sonic CD, and the expansion was also bundled with Sonic CD. It is also available for the PC as Sonic Editor. The game is considered a spiritual successor to the Genesis game Sonic 3 \& Knuckles and the Genesis port Sonic 3D Blast, as it features the same gameplay elements, including the Spin Dash, and also shares many levels and music with those games...\\

Nucleus (1.5B) &  A version of Sonic the Hedgehog was developed by Ancient and released in 1991 for Sega's 8 - bit consoles, the Master System and Game Gear. It was set in the garden city of Sky Sanctuary and centered around Sonic's new helper, the pre-historic pillar-crusher-man Waddle Dee. Sonic the Hedgehog all the way was released for Game Gear in 1992. The author of the story speaks only for himself, but here it is.

TO: Sonic the Hedgehog

FROM: Anti-Socials

SUBJECT: Argyle DeBrosse

Contents show]

What's that? You want me to tell you what I did, huh? Well, some things never change.

Yeah. Basically, you. Now listen: I saw you on the NWA charts for ages. I heard about how great you were, and that you were a great singer. And then... later, after hearing it five times in a row, that's when I realized that I couldn't win this particular battle. You and your music have the edge...\\
\bottomrule
\end{tabular}}
\caption{\label{tab:qualitative1} Qualitative example of contrastive decoding versus nucleus sampling. CD produces more coherent text both in content and style, whereas nucleus sampling produces text that suffers from topic and style drifts.
}
\end{table*}

We include a truncated qualitative example in \cref{tab:qualitative1}. The nucleus sampling output shows a topic drift from a video game to music, and part of the generated text includes the format of an email; moreover, there is a style shift from third person narrative style to first person conversational style. These features match the noisy pre-training distribution of internet data, but are not desirable in the context of this prompt. Contrastive decoding output stays on topic with the prompt and elaborates on various aspects of the game, making it more coherent in both content and style.  We include more qualitative examples in the appendix. 

\section{Ablation Studies} 

\subsection{Size of Amateur and Expert LMs}
\label{sec:ablation_size}

\begin{figure}
    \centering
     \hspace{-30pt}
    \includegraphics[width=0.25\textwidth, page=1]{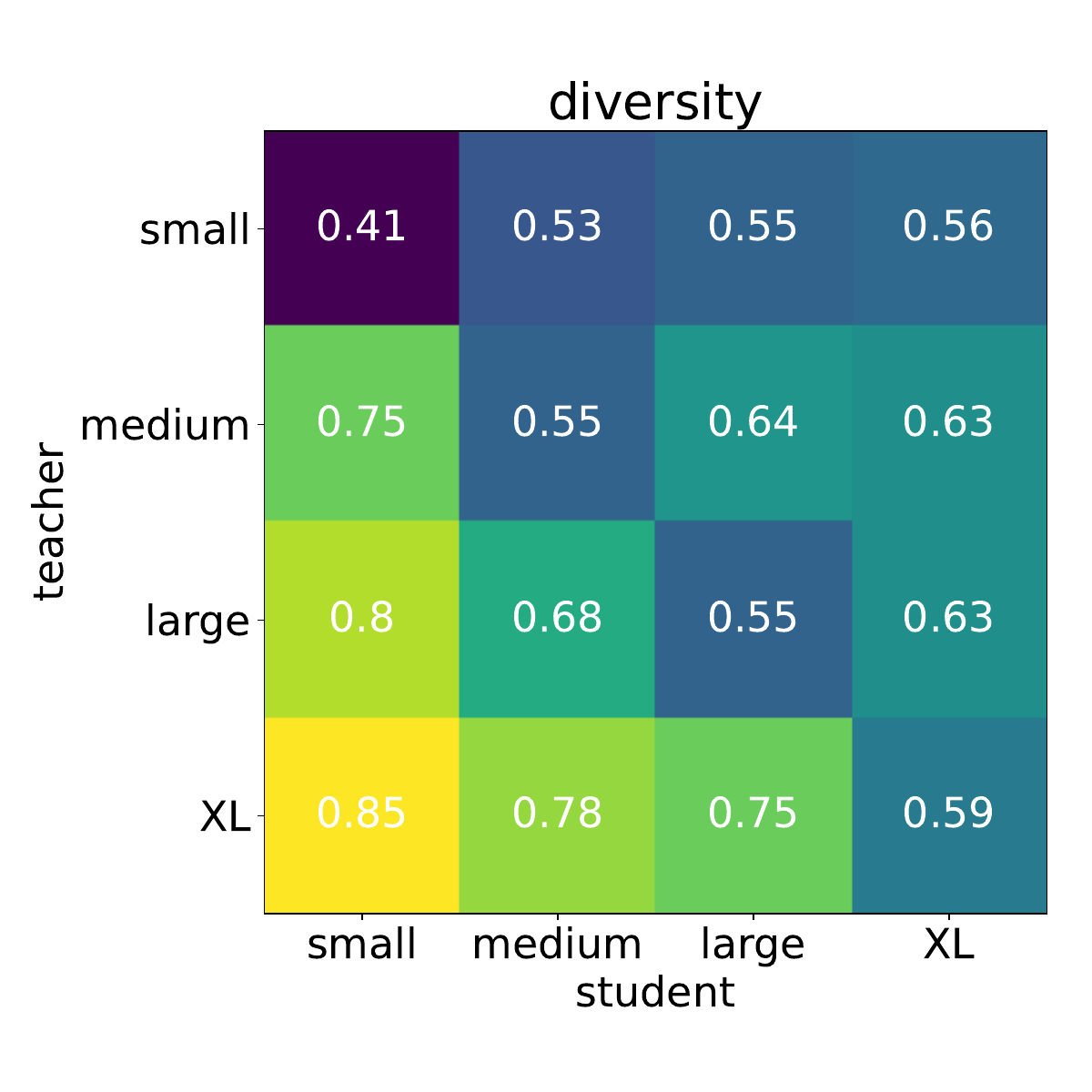} \hspace{-10pt}
    \includegraphics[width=0.25\textwidth, page=2]{figs/heatmap_gpt2.pdf}\hspace{-10pt}
     \caption{\label{fig:abla_size} Generation quality when applying contrastive decoding to expert and amateur LMs of different scales (\cref{sec:ablation_size}). We explore the expert-amateur combination within GPT-2 family (OPT family results in the appendix). 
     We find the larger scale gap between the expert and the amateur LMs, the more text quality improves. \looseness=-1
    }
\end{figure}
Recall in \cref{ssec:amateur}, we provide intuition that choosing smaller LMs as the amateur should improve contrastive decoding results. We empirically verify this in \cref{fig:abla_size}. 

The diagonal entries use the same model as expert and amateur, yielding highly repetitive text (low diversity score), because we cannot exploit any contrast between two identical LMs. The upper triangular entries use an expert LM that is smaller than the amateur LM, and this counter-intuitive setup leads to inferior text quality. The lower triangular entries use an expert LM that is larger than the amateur LM, resulting in higher quality text, as measured by both diversity and MAUVE. In particular, the optimal design is to select the largest LM as the expert and the smallest one as the amateur (lower left corner).  

Does this trend generalize to extremely low capacity LMs like n-gram models? We find that employing a trigram LM as the amateur produces low quality text with a MAUVE score of only 0.73. Our findings indicate that contrastive decoding benefits most with an amateur LM that can emphasize the failure modes of the expert LM, and the mistakes of a low-capacity n-gram model do not highlight failure modes of an expert LM.

\subsection{The Impact of Amateur Temperature} 
\label{ssec:ablahyper}
\begin{figure}
    \centering
    \includegraphics[width=0.35\textwidth, page=1]{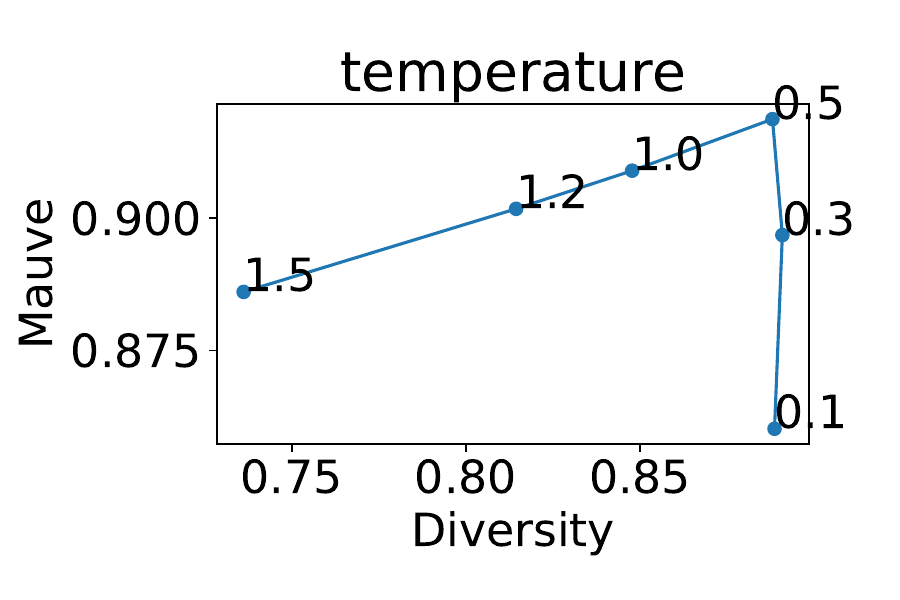}  
\caption{\label{tab:hyperparam}
Ablation studies for amateur temperature $\tau$ (\cref{ssec:ablahyper}). The figure shows how \mauve \ and diversity  score change as we vary the $\tau$ values, labeled next to each dot. 
We find that $\tau \in [0.5, 1.0]$ robustly result in high generation quality. For main results we use $\tau = 0.5$ for GPT-2 and  $\tau = 1.0$ for OPT. 
}
\end{figure}

Recall in \cref{ssec:contrastive}, we introduced the amateur LM temperature $\tau$ as a hyperparameter.  
We study how sensitive our method is to $\tau$ as shown in \cref{tab:hyperparam}.

Large $\tau$ brings the amateur distribution closer to the uniform distribution, which makes contrastive decoding generate repetitive text, as repetition is no longer penalized. Small $\tau$ makes the amateur LM more spiky and emphasizes undesired amateur behaviors, leading to better outputs from contrastive decoding. As shown in \cref{tab:hyperparam}, we find that setting $\tau$ in $[0.5, 1.5]$ attains good and robust performance in coherence and fluency. 

\subsection{Sampling v.s. Search} 
\label{ssec:abla_search_sample}

\begin{table}
\centering
\resizebox{0.4\textwidth}{!}{
\begin{tabular}{llccccccc}
& name&     \diversity &    \mauve&    \coherence & PPL  \\
\toprule
\multirow{2}{*}{\rotatebox{90}{1.5B}}
& CD (search)&  \textbf{0.89}&   \textbf{0.92}&  \textbf{0.69}& \textbf{17.77} \\
& CD (sample)&  0.81&   0.85&  0.68&  18.48  \\
\hline \hline
\multirow{3}{*}{\rotatebox{90}{1.5B}}
& CD (full)&    0.89&   \textbf{0.92}&  \textbf{0.69}& \textbf{17.77} \\
& CD (- $\vv$) & \textbf{1.0}&    0.01&   0.23& 2e5 \\
\bottomrule

\end{tabular}
}
\caption{\label{tab:ablation_automatic} Automatic evaluation for the ablation studies of search v.s. sampling the contrastive objective (\cref{ssec:abla_search_sample}) and the importance of the plausibility constraint $\vv$ (\cref{ssec:plausible}).  \looseness=-1}
\end{table}

\begin{table}
\centering
\resizebox{0.5\textwidth}{!}{
 \hspace{-10pt}
\begin{tabular}{lcc|ccc|ccccccllllll}
& & &  \multicolumn{3}{c}{coherence} & \multicolumn{3}{c}{fluency} \\
& A & B  & A &     same  &    B  &  A &     same  &    B  & \\
\toprule

1.5b & CD (search) & CD (sample) & \textbf{0.535} & 0.04 & 0.424 & \textbf{0.434} & 0.333 & 0.232 \\ 
13b &  CD (search) & CD (sample) & \textbf{0.465} & 0.162 & 0.374 & \textbf{0.475} & 0.131 & 0.394 \\ 
1.5b & CD (full) &  CD (-context) & \textbf{0.424} & 0.172 & 0.404 & \textbf{0.364} & 0.283 & 0.354 \\
\end{tabular}
}
\caption{\label{tab:ablation_human} Human evaluation for the ablation studies of search v.s. sampling the contrastive objective (\cref{ssec:abla_search_sample}) and ignoring prefix v.s. including prompt to the amateur LM (\cref{ssec:abla_prompt}). 
CD (-context) denotes the ablation experiments where we condition on the entire context for both amatuer and expert, and CD(full) conditions the amateur only on the last context token. }
\end{table}

Recall that contrastive decoding is a \emph{search}-based approach that maximizes the contrastive objective subject to plausibility constraints. We explore a \emph{sampling} alternative based on the same objective. Specifically, we normalize the $\score(x_i; x_{<i})$ (defined in \cref{ssec:contrastive}) via softmax into a probability distribution from which we sample the next token. As shown in \cref{tab:ablation_automatic} and \cref{tab:ablation_human}, we find that sampling from this objective produces lower quality text than searching under the objective. According to automatic and human evaluations, CD (sample)'s fluency and coherence rating consistently falls behind CD (search), but sampling still yields reasonably good outputs. \looseness=-1

\subsection{Plausibility Constraints} 
\label{ssec:plausible}
In \cref{ssec:plausibility}, we describe why including the feasibility constraints is critical. Here, we conduct an ablation study verifying this claim by removing the plausibility constraints $\vv$. We find that the generation outputs suffers from severe fluency issues, as easily shown by its MAUVE score of 0.01 in the CD(-$\vv$) row of \cref{tab:ablation_automatic}. 

\subsection{Prompt Inclusion} 
\label{ssec:abla_prompt}
We further experiment with ablating the prompt context on the amateur LM (\cref{ssec:amateur}), by letting the expert LM and amateur LM both condition on the entire $\xprompt$. \cref{tab:ablation_human} shows that the ablation slightly hurts coherence and fluency. 

\section{Related Work}
\paragraph{Decoding Methods.} 
Decoding algorithms can be broadly classified as either search or sampling algorithms. Current search methods (\eg greedy and beam search) attain accurate generation in goal-driven tasks (\eg summarization), but suffers from tedious and repetitive outputs in open-ended settings (\eg story generation). Current sampling methods (\eg nucleus \citep{Holtzman2020Nucleus},  top-k \citep{fan-etal-2018-hierarchical}, and typical decoding \citep{Meister-2022-typica}) produces more diverse and interesting text in open-ended settings, but suffers from unnatural topic drift. Contrastive decoding avoids topic drift by using search, and outperforms nucleus and top-k sampling in coherence while maintaining or improving fluency and lexical diversity. 

\paragraph{Contrast in Text Generation.}
The idea of contrast for text generation has been explored in diverse settings \cite{he-etal-2019-pun, li-etal-2016-diversity, Su2022ACF}. The closest work to ours is DExpert \cite{liu-etal-2021-dexperts}, which studies controllable text generation by contrasting an trained expert model (on non-toxic data) and a trained anti-expert model (on toxic data) to produce text that is non-toxic. In this work, we focus on open-ended text generation and show that it is possible to get domain- and task-agnostic anti-experts simply by using a smaller LM. 
Contrastive decoding contrasts off-the-shelf LMs of different scales to produce high quality text, without any training. 

\section{Conclusion and Future Work}
We propose contrastive decoding, a search-based decoding approach that contrasts LMs of different scales. We evaluate our approach on open-ended  text generation, and find that it improves over the prevalent methods like nucleus sampling in both fluency and coherence. 

As future work, the idea of contrasting an expert (larger LM) and an amateur (smaller LM) can be expanded to myriad setups, for instance, contrasting an early checkpoint of an LM and a later checkpoint of the LM. We hope that this paper can encourage more exploration of how to use contrasting language models.

\section*{Limitations} 
In this paper, we focus on open-ended text generation and demonstrate the effectiveness of contrastive decoding. We would like contrastive decoding to also work well for task-oriented generation settings such as summarization and machine translation. However, the idea of contrasting models across different scales (larger expert LM and smaller amateur LM) is not directly applicable, because the modes of both amateur LM and expert LM are of high quality. Empirically, having a smaller summaization model (BART-small finetuned on summarization data) as the amateur LM yields lower ROUGE score than employing a uniform distribution as the amateur LM, which is equivalent to beam search based on log-probabilities. As future work, we aim to study the necessary properties of amateur LM to empower task-oriented generation (e.g. summarization, table-to-text). 

\appendix 

\bibliography{acl2023}
\bibliographystyle{acl_natbib}

\appendix
\newpage
\section{CD-Score Analysis}
\label{app:score}
In order to emprically justify our contrastive objective, we report the likelihood scores and contrastive scores for repetitive text, reference and sampling outputs. As shown in \cref{tab:score}, we find that reference text scores highest under our contrastive loss objective, whereas the likelihood maximization objective ranks the undesired repetitive text the highest.

Averaging across the wikitext data, repetitive text receives a likelihood score of -0.79 per token, reference text receives -3.20, and sampling output receives -2.93. Contrastive objective on the other hand, assigns 0.21 to repetitive text, 0.62 to reference text, and 0.59 to sampling text. This trend is consistent with observation in the \cref{tab:score}, and contrastive scores correctly assigns highest ranking to reference text.  

\begin{table*}
\small
\centering
\resizebox{0.95\linewidth}{!}{
\begin{tabular}{p{1.5cm}p{10cm}cc}
\toprule
Source & Text & $\log \pexpert$ & $\log \pexpert - \log \pamateur$ \\ 
\midrule
Repetitive Output & Headlam served as Officer Commanding North @-@ Western Area in 1946, and as Director of Training from 1947 to 1950. In 1950 – 51, he was Commanding Officer of the 1st Battalion, 7th Infantry, 101st Airborne Division. He was awarded the Distinguished Service Cross for his actions in the Battle of the Bulge. He was awarded the Distinguished Service Medal for his actions in the Battle of the Bulge. He was awarded the Silver Star for his actions in the Battle of the Bulge. He was awarded the Bronze Star for his actions in the Battle of the Bulge. He was awarded the Purple Heart for his actions in the Battle of the Bulge. He was awarded the Distinguished Service Medal for his actions in the Battle of the Bulge. He was awarded the Silver Star for his actions in the Battle of the Bulge. He was awarded the Bronze Star for his actions in the Battle of the Bulge. He was awarded the Purple Heart for his actions in the Battle of the Bulge. He was awarded the Distinguished Service Medal for his actions in the Battle of the Bulge. He was awarded the Silver Star for his actions in the Battle of the Bulge. He was awarded the Bronze Star for his actions in the Battle of the Bulge. He was awarded the Purple Heart for his actions in the Battle of the Bulge. He was awarded the Distinguished &  \textbf{-222.1} & 56.7\\

Reference & Headlam served as Officer Commanding North @-@ Western Area in 1946, and as Director of Training from 1947 to 1950. In 1950 – 51, during the Malayan Emergency, he was stationed at Singapore as commander of No. 90 ( Composite ) Wing and, later, RAF Tengah. He twice served as acting Air Member for Personnel, in 1957 and 1959 – 60, receiving appointment as a Commander of the Order of the British Empire in 1958. Promoted air vice marshal, he successively held the positions of Air Officer Commanding ( AOC ) Operational Command in 1961 – 62, AOC No. 224 Group RAF from 1962 to 1965 during the Indonesia – Malaysia Konfrontasi, Deputy Chief of the Air Staff in 1965 – 66, and AOC Support Command in 1966 – 67. He was appointed a Companion of the Order of the Bath in 1965. Following a posting to London as Head of the Australian Joint Services Staff from 1968 to 1971, he retired from the Air Force and died in Melbourne five years later.  & -658.6 & \textbf{185.9} \\ 

Nucleus Output &  Headlam served as Officer Commanding North @-@ Western Area in 1946, and as Director of Training from 1947 to 1950. In 1950 – 51, Headlam promoted the application of the Chief of Staffs precepts, making it a common practice for subordinates to accomplish their assignments as commanded, and superiors acknowledged Headlam's brilliant leadership under fire by naming him *Unsinkable* – a title he came to regret. When Headlam retired in 1951, he had other ideas. He chose the position of athenium Depository Commander in Canada's Yukon Territory, and served there from 1951 to 1961. In 1961, he embarked on a tour of duty in Japan, where he commanded a squad of thirty-five men in the Skywolf division. On 13 December 1962 he made a daring cross-country dash across Great Bear Lake in Canada's Yukon Territory, and back to western Edmonton. Upon returning to the territory he ordered the corps commander to establish a single battalion headquarters out of the 40th Cavalry Regiment. Soon afterwards he was recalled to duty with the 50th Armored Group for duty in Dakotas, Minnesota, and within eighteen months he became a full-time consultant to the Canadian Military Command, Area GHQ. Under his leadership the 20th Armored Regiment sustained severe combat in Kenya, Tanzania, Democratic Republic of the Congo, Ethiopia and Rundu. He retired from the Canadian Armed Forces as Lieutenant & -863.1 & 158.9\\

\bottomrule
\end{tabular}}
\caption{\label{tab:score} We report the likelihood scores and contrastive scores for repetitive text, reference and sampling outputs. We find that reference text scores highest under our contrastive loss objective, whereas the likelihood maximization objective ranks the undesired repetitive text the highest. }
\end{table*}

\section{Quantitative Analysis of LM decoding} 
The pre-trained LMs are flawed in both coherence and repetition, and they make similar mistakes regardless of the sizes: for maxprob decoding, the 4-gram repeat rate is 71\% for GPT-2 XL, and 40\% for GPT-3 Davinci (both are unacceptably high). For sampling, the coherence score is 0.56 for GPT-2 XL and 0.57 for GPT-3 Davinci (both are lower than GPT-2 XL's CD results of 0.69). 

\section{CD as Distinguishability objective}

Recall from \cref{ssec:contrastive}, our objective $\log \frac{\pexpert(\xcont \mid \xprompt)}{\pamateur(\xcont \mid \xprompt)}$ can intuitively be interpreted as factoring out amateur tendencies from the expert LM. Formally, the argmax $\xcont$ of our contrastive objective also maximizes the pointwise mutual information $\PMI(\xcont, \II = 1)$, where $\II$ is an indicator variable that determines the source of generated text: $\II = 1$ for text generated by the expert and $\II = 0$ for text generated by the amateur.
\begin{align*}
& \PMI(\xcont, I=1) = \log \frac{p(\xcont | I=1)}{p(\xcont)} \\ 
& = \log \frac{\pexpert(\xcont)}{0.5 \pexpert(\xcont) + 0.5 \pamateur(\xcont)}\\
& = -\log ( 0.5 + 0.5 \frac{\pamateur(\xcont)}{\pexpert(\xcont)}),
\end{align*}
This leads to a formal interpretation of our objective: it favors text that has high PMI with the indicator variable $I=1$, i.e., the most distinguishable text as having originated from the expert LM, rather than the amateur LM. 

\section{Additional Related Work}

\paragraph{Training Methods.}
Prior works often aim to improve text generation quality by further training a given LM. A common approach is to fine-tune the LMs on domain specific data, which improves the relevance of generated text, but fails to fundamentally address fluency or coherence problems \cite{DeLucia-2020-decoding}. To tackle these model specific issues, many works craft novel training objectives. For example unlikelihood training \cite{Welleck2020Neural} explicitly penalizes repetition; contrastive training \cite{Su2022ACF} separates out the LM hidden states to boost diversity. Furthermore, many methods alleviate exposure bias by combining teacher-forcing and student-forcing at training time \cite{Lamb-2016-professor, Venkatraman_Hebert_Bagnell_2015,Ranzato-2016-sequence,wiseman-rush-2016-sequence}. 
Despite the effectiveness of these approaches, they require training model parameters on these crafted objectives, which can be prohibitively expensive for ever-larger models. In contrast, our method uses frozen LMs and requires no training. We simply take off-the-shelf pre-trained language models of different sizes, and exploit their differences to improve text generation quality. 

\paragraph{Contrast in Text Generation.}
The idea of contrast for text generation has been explored in diverse settings. 
In pun generation, \newcite{he-etal-2019-pun} contrasts the same LM with global versus local context to select tokens that are plausible globally but surprising locally. In dialog generation, \newcite{li-etal-2016-diversity} contrasts the same dialog model with and without preceding chat history in order to generate relevant responses. \newcite{Su2022ACF} fine-tuned language models on a contrastive training objective to separate token representations, which in turn improves generation diversity. 

The closest work to ours is DExpert \cite{liu-etal-2021-dexperts}, which studies controllable text generation by contrasting an trained expert model (on non-toxic data) and a trained anti-expert model (on toxic data) to produce text that is non-toxic. In this work, we focus on open-ended text generation and show that it is possible to get domain- and task-agnostic anti-experts simply by using a smaller LM. Contrastive decoding uses the observation that smaller LMs are more susceptible to the undesirable behaviors, and contrasts off-the-shelf LMs of different scales to produce high quality text, without any training. 

\section{Potential Ethics Risks and Societal Impact} 
Contrastive decoding aims to produce fluent and coherent continuation of a given prompt. However, as the generation quality improves, one can imagine more powerful disinformation (e.g., automatic generation of fake news) that are hard to distinguish from human written text. Towards this end, it might be worth augmenting current decoding techniques to also watermark the generated outputs without affecting its quality. 

\section{Compute Resources} 
We use NVIDIA RTX A5000 and A100 GPU to run the decoding experiments. All the decoding is done by one GPU. For OPT-13b, we use fp16 to reduce the required amount of GPU memories. CD generates one continuation of length 256 tokens (with batchsize of 1) in 8 seconds on NVIDIA RTX A5000. 

\section{Human Evaluation Details} 
\begin{figure*}
    \centering
    \includegraphics[width=0.8\textwidth,]{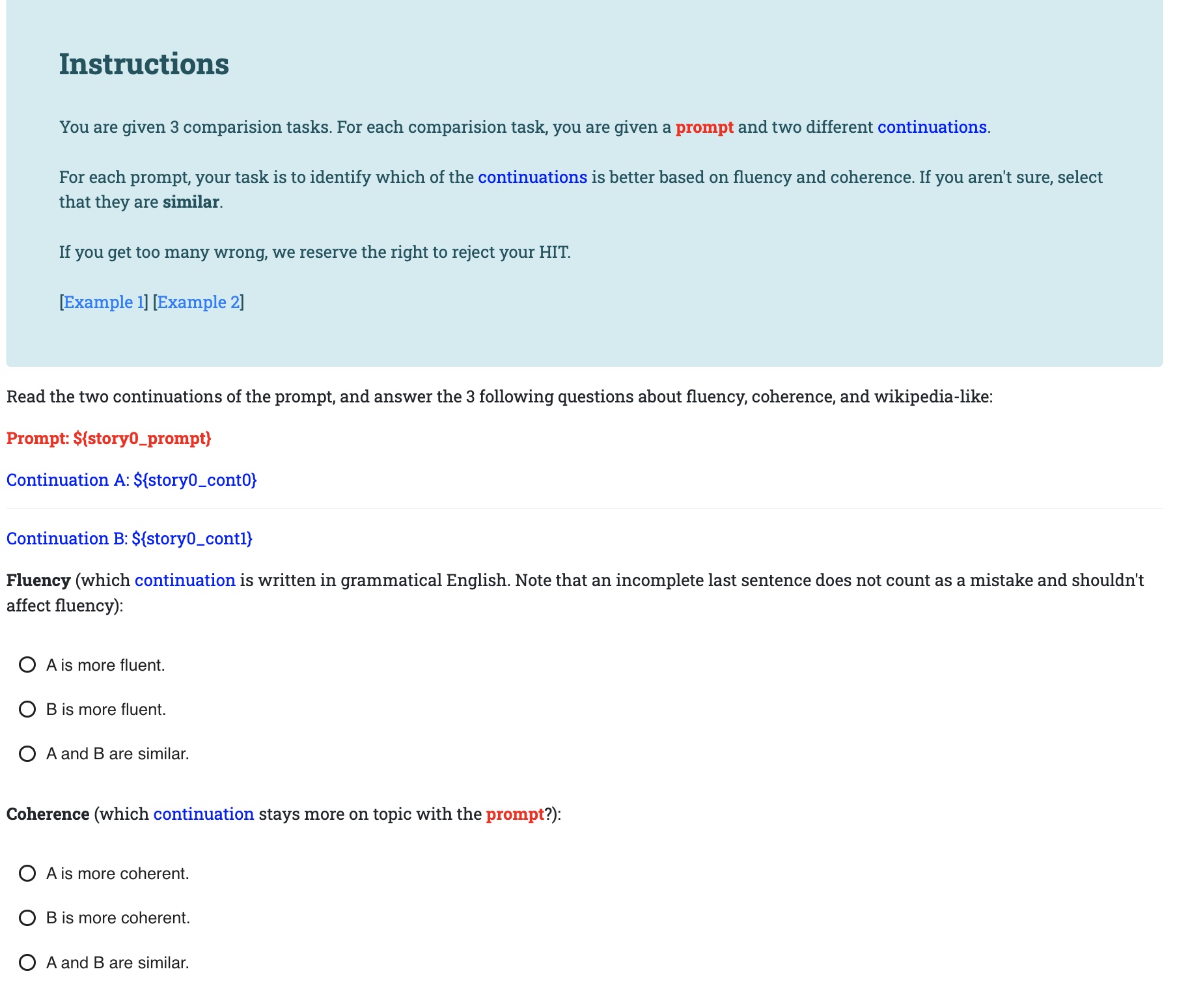}
    \caption{Human evaluation instructions and interface we post to Amazon Mechanical Turk platform.}
    \label{fig:mturk_interface}
\end{figure*}

We report the instruction given to the Amazon mechanical turkers in \cref{fig:mturk_interface}, and we explain the annotation results will be used towards distinguishing text generation qualities. 

We conduct a pre-qualification round of 60 people to ensure the participants understand the task and are capable of judging fluency and coherence, resulting in around 20 people qualified. 

We assign 20 minutes to each HITs, which consists of three comparison tasks. Each HITs takes 14 minutes on average to complete. We pay \$4.5 for each HITs, which adds up to an hourly payment of  \$18, which is adequate given the participants' demographic. Our human evaluation project received approval from the ethics review. 

\section{Expert and Amateurs from Different model Families}
In the main paper, we focus in the settings where the experts and the amateurs come from the same model family (e.g., GPT-2 small v.s. GPT-2 XL; OPT-125M v.s. OPT-13B), because the tokenizer is the same within each model family. However, contrastive decoding still works when the expert and amateur models come from different model families. In particular, we use GPT-J as the expert and GPT-2 small as the amateur (the two models are pre-trained on different datasets by different companies, but share the same tokenizer). We find that CD yields \mauve=0.93, \diversity=0.91, which is better than GPT-2 XL’s CD results. 
\section{Full Automatic Evaluation Results} 
\label{app:full}
In \cref{tab:automatic_eval}, we report diversity, MAUVE, and $\coherence$. In the tables (\cref{tab:wikitext} for wikitext, \cref{tab:wikinews} for wikinews,  \cref{tab:story} for story), we also include $\repn$ metrics for $n=2,3,4$ and perplexity (PPL)  under GTP-2 medium, along with MAUVE, $\coherence$ and $\diversity$.

\begin{table*}[!]
\centering
\begin{tabular}{lllllllllllllllll}
& name&                                          rep-2&  rep-3&  rep-4&  div&    mau&    co& PPL & \\
\toprule
\multirow{7}{*}{\rotatebox{90}{OPT-6.7B}}

& greedy&     71.95&  68.34&  65.98&  0.03&   0.07&   0.63&   5.2&\\
& k-50&       9.32&   3.79&   2.48&   0.85&   0.86&   0.61&   23.55&\\
& k-10&       16.18&  8.34&   5.77&   0.72&   0.73&   0.64&   15.33&\\
& p-0.95&     7.71&   3.3&    2.31&   0.87&   0.85&   0.59&   32.2&\\
& typical-0.95&       5.02&   1.62&   1.02&   0.92&   0.89&   0.56&   50.73&\\
& CD-1.0&     8.68&   2.09&   0.65&   0.89&   0.91&   0.69&   29.71&\\

\midrule
\multirow{7}{*}{\rotatebox{90}{OPT-13B}} 
& greedy&      71.52&  67.88&  65.53&  0.03&   0.08&   0.63&   5.37&\\
& k-10&        15.81&  8.38&   6.02&   0.72&   0.77&   0.64&   15.73&\\
& k-50&        9.06&   3.76&   2.54&   0.85&   0.83&   0.61&   23.88&\\
& typical-0.95&        5.09&   1.84&   1.27&   0.92&   0.89&   0.55&   50.67&\\
& p-0.95&      6.96&   2.74&   1.85&   0.89&   0.86&   0.58&   33.01&\\
& CD-1.0&      7.55&   1.63&   0.47&   0.91&   0.91&   0.69&   32.53&\\

\midrule
\multirow{7}{*}{\rotatebox{90}{GPT2-XL}} 
& k-50&        8.24&   2.92&   1.78&   0.87&   0.79&   0.61&   19.96&\\
& p-0.95&      5.25&   1.68&   1.07&   0.92&   0.87&   0.57&   34.35&\\
& typical-0.95&        3.59&   1.01&   0.65&   0.95&   0.84&   0.53&   57.8&\\
& greedy&      76.3&   73.58&  71.8&   0.02&   0.05&   0.62&   4.19&\\
& k-10&        15.45&  7.47&   4.95&   0.74&   0.76&   0.64&   12.81&\\
& CD-1.0 &       9.19&   1.81&   0.41&   0.89&   0.92&   0.69&   24.66&\\
& beamprefix-0.8&      6.88&   1.19&   0.24&   0.92&   0.9&    0.7&    24.46&\\

\end{tabular}
\caption{\label{tab:wikitext} Automatic evaluation results for wikitext. }
\end{table*}

\begin{table*}[]
\centering
\begin{tabular}{lllllllllllllllll}
& name&                                          rep-2&  rep-3&  rep-4&  div&    mau&    co& PPL & \\
\toprule
\multirow{7}{*}{\rotatebox{90}{OPT-6.7B}}
& greedy&                     61.44&  57.94&  56.06&  0.07&   0.26&   0.65&   6.45&\\
& k=50&                       6.17&   2.07&   1.27&   0.91&   0.92&   0.64&   19.99&\\
& k=10&                       9.48&   3.89&   2.47&   0.85&   0.88&   0.67&   14.05&\\
& p=0.95&                     5.65&   1.96&   1.27&   0.91&   0.92&   0.62&   22.89&\\
& typical=0.95&               4.19&   1.23&   0.77&   0.94&   0.93&   0.58&   34.11&\\
& CD-1.0&     5.62&   1.19&   0.37&   0.93&   0.95&   0.69&   25.42&\\
\midrule
\multirow{7}{*}{\rotatebox{90}{OPT-13B}} 
& greedy&                      59.51&  55.84&  53.9&   0.08&   0.3&    0.65&   7.05&\\
& k-50&                        6.03&   1.95&   1.18&   0.91&   0.92&   0.64&   20.29&\\
& k-10&                        9.06&   3.49&   2.1&    0.86&   0.9&    0.66&   14.34&\\
& p-0.95&                      5.21&   1.54&   0.9&    0.92&   0.92&   0.62&   22.77&\\
& typical=0.95&                4.17&   1.23&   0.77&   0.94&   0.9&    0.59&   33.63&\\
& CD-1.0&      5.27&   1.03&   0.26&   0.94&   0.94&   0.69&   27.24&\\
\midrule
\multirow{7}{*}{\rotatebox{90}{GPT2-XL}} 
& greedy&                      69.55&  66.68&  65.0&   0.04&   0.14&   0.65&   4.48&\\
& k-50&                        5.95&   1.73&   0.93&   0.92&   0.88&   0.64&   16.35&\\
& k-10&                        10.14&  4.06&   2.46&   0.84&   0.86&   0.66&   11.04&\\
& p-0.95&                      4.62&   1.24&   0.7&    0.94&   0.9&    0.6&    22.32&\\
& typical-0.95&                3.4&    0.82&   0.46&   0.95&   0.91&   0.56&   35.35&\\
& beamprefix-0.8&              4.8&    0.76&   0.14&   0.94&   0.94&   0.7&    20.02&\\
& CD-1.0 &               6.7&    1.19&   0.24&   0.92&   0.94&   0.69&   21.59&\\
\end{tabular}
\caption{\label{tab:wikinews} Automatic evaluation results for Wikinews dataset. }
\end{table*}

\begin{table*}[]
\centering
\begin{tabular}{lllllllllllllllll}
& name&                                          rep-2&  rep-3&  rep-4&  div&    mau&    co& PPL & \\
\toprule
\multirow{7}{*}{\rotatebox{90}{OPT-6.7B}}
& k-10&                   13.53&  5.96&   3.64&   0.78&   0.89&   0.54&   14.15&\\
& k-50&                   6.66&   2.01&   1.05&   0.91&   0.9&    0.51&   22.48&\\
& greedy&                 77.86&  75.01&  73.04&  0.01&   0.05&   0.51&   4.93&\\
& p-0.95&                 5.12&   1.47&   0.82&   0.93&   0.9&    0.48&   30.71&\\
& typical-0.95&           3.73&   0.95&   0.55&   0.95&   0.89&   0.45&   47.56&\\
& CD-1.0& 9.52&   2.67&   1.03&   0.87&   0.94&   0.61&   22.64&\\
\midrule
\multirow{7}{*}{\rotatebox{90}{OPT-13B}} 
& greedy&                  76.37&  73.22&  71.03&  0.02&   0.05&   0.51&   5.1&\\
& typical-0.95&            3.65&   0.84&   0.43&   0.95&   0.91&   0.46&   47.41&\\
& k-10&                    12.91&  5.55&   3.31&   0.8&    0.87&   0.54&   14.42&\\
& k-50&                    6.57&   1.95&   1.03&   0.91&   0.9&    0.51&   22.47&\\
& p-0.95&                  4.97&   1.32&   0.7&    0.93&   0.91&   0.48&   31.05&\\
& CD-1.0&  8.56&   2.15&   0.76&   0.89&   0.94&   0.62&   23.95&\\
\midrule
\multirow{7}{*}{\rotatebox{90}{GPT2-XL}} 
& p-0.95&                  4.27&   0.9&    0.39&   0.94&   0.91&   0.46&   30.47&\\
& k-50&                    6.52&   1.68&   0.75&   0.91&   0.87&   0.51&   18.83&\\
& typical-0.95&            2.95&   0.54&   0.24&   0.96&   0.88&   0.43&   49.9&\\
& k-10&                    13.57&  5.49&   2.97&   0.79&   0.81&   0.54&   11.76&\\
& greedy&                  81.51&  79.2&   77.53&  0.01&   0.03&   0.49&   3.06&\\
& CD-1.0&           12.8&   3.68&   1.25&   0.83&   0.94&   0.64&   16.36&\\
& beamprefix-0.8&          7.71&   1.33&   0.27&   0.91&   0.9&    0.63&   18.37&\\

\end{tabular}
\caption{\label{tab:story} Automatic evaluation results for story generation}
\end{table*}

\section{Additional Ablation Results} 
\begin{figure*}
\begin{minipage}{0.7\columnwidth}
    \centering
    \includegraphics[width=1.0\textwidth, page=1]{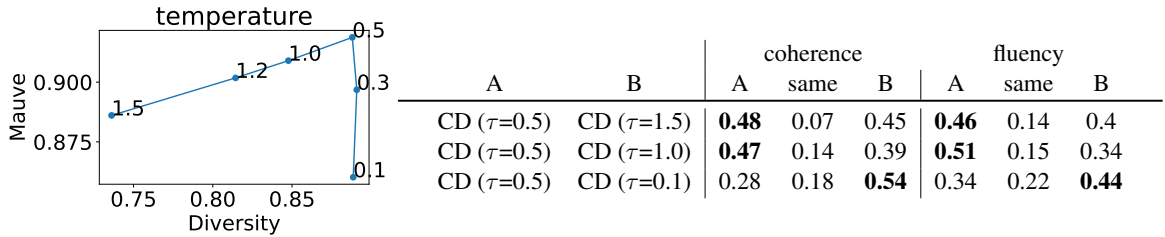}  
    \hspace{-10pt}
\end{minipage}
\begin{minipage}{1.3\columnwidth}
\resizebox{1.0\textwidth}{!}{
 \hspace{-10pt}
\begin{tabular}{lcc|ccc|ccccccllllll}
& & &  \multicolumn{3}{c}{coherence} & \multicolumn{3}{c}{fluency} \\
& A & B  & A &     same  &    B  &  A &     same  &    B  & \\
\toprule

& CD ($\tau$=0.5) &  CD ($\tau$=1.5) & \textbf{0.48} & 0.07 & 0.45 & \textbf{0.46} & 0.14 & 0.4 \\
& CD ($\tau$=0.5) & CD ($\tau$=1.0) & \textbf{0.47} & 0.14 & 0.39 & \textbf{0.51} &  0.15 & 0.34 \\ 
& CD ($\tau$=0.5) & CD ($\tau$=0.1) & 0.28 & 0.18 & \textbf{0.54} & 0.34 & 0.22 & \textbf{0.44} \\ 
\end{tabular}
}
\end{minipage}
\caption{\label{apptab:temperature}
Ablation studies for CD's sensitivity to amateur temperature $\tau$ (\cref{ssec:ablahyper}). The left plot is based on automatic metrics, and it shows how MAUVE and diversity score change as we vary the $\tau$ values, labeled next to each dot. The right table is based on human evaluation, and we report coherence and fluency preference in the same format as \cref{tab:human_eval}.
We find that $\tau \in [0.5, 1.0]$ robustly result in high generation quality. For main results we use $\tau = 0.5$ for GPT-2 and  $\tau = 1.0$ for OPT. }
\end{figure*}

As shown in \cref{apptab:temperature}, we report additional results for the ablation study of amateur temperature. We find that $\tau \in [0.5, 1.0]$ robustly result in high generation quality.

\begin{figure*}
    \centering
     \hspace{-30pt}
    \includegraphics[width=0.25\textwidth, page=1]{figs/heatmap_gpt2.pdf} \hspace{-10pt}
    \includegraphics[width=0.25\textwidth, page=2]{figs/heatmap_gpt2.pdf}\hspace{-10pt}
    \includegraphics[width=0.25\textwidth, page=1]{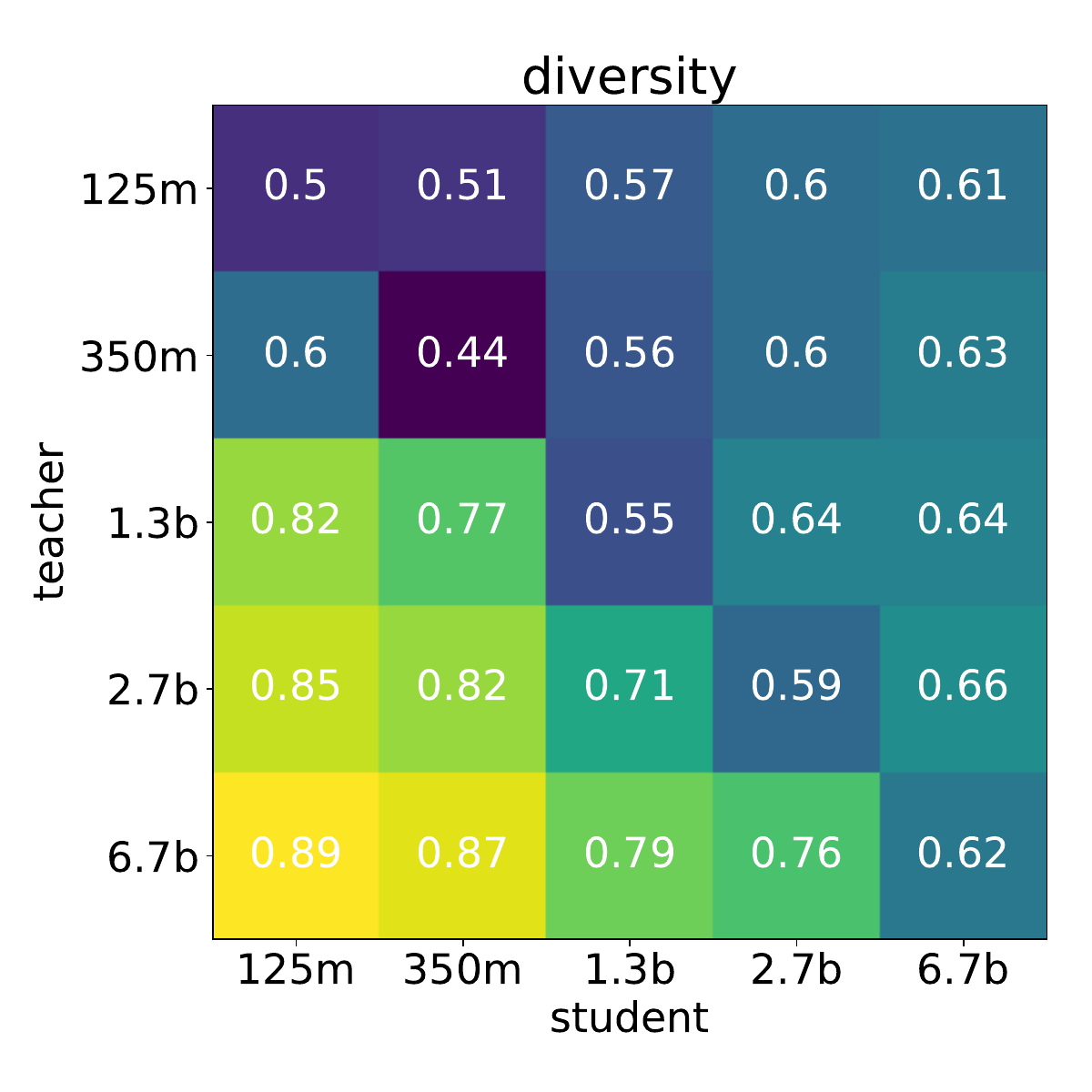} \hspace{-10pt}
     \includegraphics[width=0.25\textwidth, page=2]{figs/heatmap_opt.pdf} \hspace{-30pt}
     \caption{\label{appfig:abla_size} Generation quality when applying contrastive decoding to expert and amateur LMs of different scales (\cref{sec:ablation_size}). The left two plots explore the expert-amateur combination within GPT-2 family. 
     The right two plots explore size combination in the OPT family. 
     We find that within the same LM family, the larger scale gap between the expert LM versus the amateur LM, the more text quality improves.
    }
\end{figure*}

In \cref{appfig:abla_size}, we provide additional results on the amateur-expert size combinations for the OPT family and GPT-2 family. We find that within the same LM family, the larger scale gap between the expert LM versus the amateur LM, the more text quality improves.

\section{Additional Ablation Results for Sample v.s. Search}
Recall in \cref{ssec:abla_search_sample}, we compare  sampling CD objective and searching CD objective. Here, we include extra results in  \cref{tab_app:ablation_sample_search}. We find that CD (search) outperform CD (sample) consistently across three domains and three model sizes.

\begin{table*}[]
\centering
\resizebox{1.0\textwidth}{!}{
\begin{tabular}{llccc|ccc|cccllllll}
& & \multicolumn{3}{c}{wikinews} & \multicolumn{3}{c}{wikitext} & \multicolumn{3}{c}{story} \\
& name&     div &    mauve&    coh  & div & mauve & coh  &  div &    mauve&    coh & \\
\toprule
\multirow{3}{*}{\rotatebox{90}{6.7B}}
& p=0.95&                      0.91&   0.92&   0.62 &   0.87&   0.85&   0.59 & 0.93&   0.9&    0.48&\\
& CD (search)&      0.93&   0.95&  0.69 & 0.89&   0.91&   0.69 & 0.87&   0.94&   0.61& \\
& CD (sample)&      0.86&   0.91&   0.69 & 0.79&   0.82&   0.68&  0.85&   0.93&   0.58&  \\
\midrule
\multirow{3}{*}{\rotatebox{90}{13B}} 
& p=0.95&                       0.92&   0.92&   0.62 & 0.92&   0.89&   0.55  & 0.93&   0.91&   0.48& \\
& CD (search)&  0.94&   0.94&  0.69 &  0.91&   0.91&   0.69 &  0.89&  0.94&   0.62& \\
& CD (sample) & 0.87&   0.9&    0.68 & 0.8&    0.84&   0.67 & 0.85&   0.91&   0.59& \\
\midrule
\multirow{4}{*}{\rotatebox{90}{1.5B}} 
& p=0.95&                   0.94&   0.9&    0.6&  0.92&   0.87&   0.57  & 0.94&   0.91&   0.46&\\
& CD (search) &         0.92&   0.94&   0.69 & 0.89&   0.92&   0.69 & 0.83&   0.94&   0.64 & \\
& CD (sample)&     0.88&   0.89&   0.69&   0.81&   0.85&   0.68&  0.9&    0.92&   0.59& \\
\midrule
& full CD &         0.92&   0.94&   0.69  & 0.89&  0.92&   0.69 & 0.83&   0.94&  0.64&\\

& Context Ablation&     0.92&   0.93&   0.63&  0.88&   0.88&   0.6&  0.88&   0.9&    0.53&  \\
& $\vv$ Ablation &   0.92&   0.93&   0.63&  0.88&   0.88&   0.6   & 0.88&   0.9&    0.53&  \\
\end{tabular}
}
\caption{\label{tab_app:ablation_sample_search} Full ablation studies results comparing CD (sample) and CD search \cref{ssec:abla_search_sample}. We find that CD (search) outperform CD (sample) consistently across three domains and three model sizes.  }
\end{table*}

\section{More Qualitative Examples}
We include 6 randomly sampled qualitative examples in Table \ref{tab:qua2} -- \ref{tab:qua7}.   

\begin{table*}
\small
\centering
\begin{tabular}{lp{13cm}}
\toprule
Prompt &  The 40 - metre - long ( 131 ft 3 in ) hangar was 20 metres ( 65 ft 7 in ) wide forward and 11 metres\\ 
\midrule
CD (1.5b) &  The 40 - metre - long ( 131 ft 3 in ) hangar was 20 metres ( 65 ft 7 in ) wide forward and 11 metres wide ( 33 ft 6in) aft. The hangar had two doors on the port and starboard sides and a single door in front. The
port door opened to a large, open hangar deck which extended forward to about 20 metres ( 65ft 7 in ) in width, then curved around and closed on a small, enclosed platform which had two small, circular, windows in
it, each about 10 feet by 6 feet ( 3m x 1.3 m ).
The port and forward door opened onto the forward part ( port side) of a hangar which extended to a platform on which was suspended the main wing assembly, a single vertical tailplane and two vertical fin assemblie
s, each with two horizontal planes and two horizontal stabilizer planes, all suspended by cables from a central vertical stabilizing strut. The platform had two small windows on it and a small door in it, which cou
ld also open. The platform was suspended from a large vertical strut which ran the entire width and which could also swing out to form a small platform for a single plane, or could be swung into position for the en
tire assembly to rest upon. The hangar deck extended to the rear, where it curved around to close on the small platform which had the single, small, round, window. This small platform\\

Nucleus (1.5b) &  The 40 - metre - long ( 131 ft 3 in ) hangar was 20 metres ( 65 ft 7 in ) wide forward and 11 metres ( 32 ft 3 in ) wide aft of the posvailar aircraft

Long house tractor OHP (on ramp) for refuelling of aircraft

Multi-armed bus for each aircraft

Type 60 1012 with full size service ship kitted out to have a toilet on the U-shaped port side

Alternating twin Mk-III main guns on the port side could each be fitted with fuzes to fight various targets

Interior of two cabin cabins aboard YK-SATF

Modern sleeping quarters aboard YK-SATF with internal bunk room, practically designed for two officers and 128 people (51 passengers, 193 space on D-handlers)

Room to move and all-round living facilities on U-shaped bar stand on the port side (covering port side)

Elevator to the Laundry Arcade on the port side and overhead hoist to the garage with parking (drivers) and service bus

Medium weight serial 717-5005 complete with enclosed treadle trailer and winch, bought in BOMBARDIER'S FORGOT LANDING

Modern passenger transporter and ( Grumman HUU-21 ) with new advance Z \\

CD (13b) & The 40 - metre - long ( 131 ft 3 in ) hangar was 20 metres ( 65 ft 7 in ) wide forward and 11 metres wide aft ( 34 ft 7 in ). The hangar had two doors on the port and starboard sides and a single door in the centre, which could be closed to form a sealed room. It could accommodate a Boeing B-17F Flying Fortress bomber. The aircraft were parked nose to tail, with the nose of one aircraft abeam of the centreline of the other aircraft, with their wings touching. The aircraft were parked in this fashion to prevent them colliding in flight, which was a serious hazard in the days of wooden aircraft. The hangar doors could be opened in flight, but were normally left closed. The hangar had an internal height of 8 metres ( 26 ft ) and an internal volume of 5,500 cubic feet ( 152 m 3 ).\\

Nucleus (13b) & The 40 - metre - long ( 131 ft 3 in ) hangar was 20 metres ( 65 ft 7 in ) wide forward and 11 metres ( 36 ft 10 in ). At the rear of each side of the hangar, there was a covered gangway, for walking up and down. This went from the winch room on the starboard side and descended over the steel deck until reaching the 2 x 10 sheet metal Whitworth Type 328's head. The starboard side of the hangar swung 90 degrees, towards the bridge, and ended with a 10 metre ( 33 ft ) long steel skid flat at the rear which neatly incorporated the lockers for the G-Turn seats. Total hangar deck length was 183 metres ( 603 ft ). Above the single 40m x 10m hangar, the hanger deck’s Foxtrot and RMS sections wereEach section had a length of 5.0 metres ( 16 ft 3 in ) a height of 2.6 metres ( 8 foot ) and a width of 2.7 metres ( 7 ft 2 in ). The engine room was A 40m x 10m hangar built in 1937 had a longer engine hatch to both forward ends of the deep narrow profile of the FOxtrot section, so that when doors were closed supplemented the long afterbonnet doors.The crawlway from FOxtrot to RMS halved the length of the engine room.The\\
\bottomrule
\end{tabular}
\caption{\label{tab:qua1} Qualitative Example 1.}
\end{table*}

\begin{table*}
\centering
\small
\begin{CJK*}{UTF8}{gbsn}
\begin{tabular}{lp{13cm}}
\toprule
Prompt &  In some images, Hu employed a blind embossing technique ( known as " embossed designs " ( gonghua, 拱花 \\

CD (1.5b) &  In some images, Hu employed a blind embossing technique ( known as " embossed designs " ( gonghua, {拱花}) ), in others he employed an ink brush ( gonghuai, 拱花貓 ), in yet other images he employed both techniques
, or used neither technique. The technique used to emboss Hu's characters was not known to me. The images below were all created by Hu in the late Qing Dynasty, around 1900-1900 C.E.. The first three are all from a book titled " The Book on the Great Learning ", and were all created around the year 1900-1905.

The last image, which was made sometime around the early 1920's is from an unknown work. The book in which these images are contained was published sometime around 1920-1920. The images are all in the Hu style, which was a style of calligraphic Chinese script used from around the mid 19 th Century until around 1900. Hu was the most famous and prolific of calligraphy painters of this style. The style is characterized by a strong and distinct character set and a strong, distinct line-width, and was used to write many famous Chinese poems and poems by famous Chinese poets such as Li Bai ( 杨瑜 ), and Zhuangzi (  Zhūngzì, 老子\\
Nucleus (1.5b) & In some images, Hu employed a blind embossing technique ( known as " embossed designs " ( gonghua, 拱花); In which images are embossed by straight red lines on the front and back of the paper, and the paper is then pasted to the frame. There are another method in which the paper is pinned down or carried over the back of the paper to the base of the image. Chinese calligraphy artists often used this " embossed poster, ", a style developed by my great-grandfather Tang Jian Zhong Jin ( 古尾風 )....grandson Hu Yu arrived on the scene in the year 1600....unveiled the Hu's....Image 河土喜发 Eight sections 趙別字...Early Chinese of Li Cheng Hang

Copyright  2004 by Luis Estevez

This Chinese Character A-Z Chinese Character Alphabet to Practice Your Chinese Learn Chinese Characters here to Practice with your Non-Chinese Friend. Step 1

Learn and practice the characters, Step 2

Read about the Chinese characters and...Chinese characters that are used in Chinese,...Step 3

Learn and practice the written forms of characters that have their...Chinese character characters of the modern day, Exercise 以報Learn and practice the characters, Learn and practice the \\ 

CD (13b) & In some images, Hu employed a blind embossing technique ( known as " embossed designs " ( gonghua, 拱花絲 ), where a design was embossed onto a surface and then covered by another material, usually paper or si
lk, which would reveal the design when the covering material was peeled away. In other images, Hu employed an ink wash technique ( guancha, 黃档拱花 ), which involved painting designs with ink onto silk, which was
then folded over and pressed to reveal the ink designs when the folded silk was opened. In both techniques, Hu's designs often incorporated Chinese characters, flowers, birds, or other images, often in a floral arr
angement or other geometric pattern. The designs were often repeated in different colors or in different arrangements to create an overall pattern or composition. The embossed and ink wash paintings were then often
 mounted together on a silk cloth or paper to create a scroll painting or scroll painting panel ( jiaochaoren, 街上藝術篆 ), or were framed to make a wall scroll ( xuanhua, 序印 ), which were often given as gifts o
r sold to raise funds for Hu's travels or other expenses. Hu often signed the scrolls he produced, but his signatures are often difficult or impossible to read, as they have been rubbed away over \\

Nucleus (13b) &  In some images, Hu employed a blind embossing technique ( known as " embossed designs " ( gonghua, 拱花,
覆盖恢, 2008, Scale-packaging Checkpoint (Luntai)), which enables the smoothing out of its distinct design features.
The result becomes a multi-level one of its staggering packaging.
Meanwhile, in order to turn the expected carrier type into a visual one, he designed a unique structural design in order to have multi hiding spaces for its series of Brown Helpful Products( Thanks-Sweet Brown) wit
h innovative distribution of compressed air
between them.

The manufacturer was originally originally known as this address:

Of course some users are: enter you read how to play or watch this game, please feel free to
ask, please, tell some friends did you find the site in google search. Thank You. Por favor, leas cualquier otro tema. Gracias, recio los comentarios. You can leave a comment. Thank You. Leave a comment.

CONTACTS

less surprisingly, cutting produced a very good contrast between its high quality and its price. To refer to three of his early works: Bao Bao, Nanji Yaem, Ganzi Schneider. A Lexicon of Military Land Art. Globorum
Logos: avant-garde\\
\bottomrule
\end{tabular}
\end{CJK*}
\caption{\label{tab:qua2} Qualitative Example 2.}
\end{table*}

\begin{table*}
\small 
    \centering
    \begin{tabular}{lp{13cm}}
\toprule
Prompt &   A version of Sonic the Hedgehog was developed by Ancient and released in 1991 for Sega's 8 - bit consoles, the Master System and Game\\ 
\midrule
CD (1.5b) &   A version of Sonic the Hedgehog was developed by Ancient and released in 1991 for Sega's 8 - bit consoles, the Master System and Game Gear. It featured a new level editor called the Sonic Editor and was released as an expansion pack to Sonic CD in 1992, which was itself a port from the Genesis version. The editor allowed the creation and playtesting of custom levels for Sonic CD, and the expansion was also bundled with Sonic CD. It is also available for the PC as Sonic Editor. The game is considered a spiritual successor to the Genesis game Sonic 3 \& Knuckles and the Genesis port Sonic 3D Blast, as it features the same gameplay elements, including the Spin Dash, and also shares many levels and music with those games, though it was developed by a completely new team and has no connection with either of them, and was not intended as an official sequel to either game. It also shares some level design with Sonic Adventure and Sonic Adventure DX, but not Sonic CD or Sonic CD Special Delivery, as those were not developed by Ancient at all, but by a separate developer called Team Sonic, and are considered separate titles in the franchise, with their own storylines, gameplay and characters, though the two share some level designs and some characters, and Sonic CD was the only Sonic title released for both the Master System, Game Gear, Sega MegaDrive/Genesis and PC platforms at the time, so the two\\

Nucleus (1.5b) &  A version of Sonic the Hedgehog was developed by Ancient and released in 1991 for Sega's 8 - bit consoles, the Master System and Game Gear. It was set in the garden city of Sky Sanctuary and centered around Sonic's new helper, the pre-historic pillar-crusher-man Waddle Dee. Sonic the Hedgehog all the way was released for Game Gear in 1992. The author of the story speaks only for himself, but here it is.

TO: Sonic the Hedgehog

FROM: Anti-Socials

SUBJECT: Argyle DeBrosse

Contents show]

What's that? You want me to tell you what I did, huh? Well, some things never change.

Yeah. Basically, you. Now listen: I saw you on the NWA charts for ages. I heard about how great you were, and that you were a great singer. And then... later, after hearing it five times in a row, that's when I realized that I couldn't win this particular battle. You and your music have the edge.

I dunno what got into you. But I can tell you I don't appreciate it.

I heard you on the citywide broadcasts every night when they played music. And I heard what you said about your bi-coastal attitude toward Los Angeles, and it's funny\\

CD (13b) &  A version of Sonic the Hedgehog was developed by Ancient and released in 1991 for Sega's 8 - bit consoles, the Master System and Game Gear. It featured the original character design of Naoto Ohshima, who also designed Sonic for Sega's 16-bit consoles, the Mega Drive/Genesis and Game Gear. The game's title screen features Sonic, Tails and Knuckles running in place on a background of a hedge maze. The title screen music is "Hedgehog Hill", a track from Sonic CD composed by Yuji Naka and Masaya Matsuura, which also plays when the player first enters a stage in Sonic 2 and when Sonic runs up a slope in the first Sonic 3D Blast level, and also appears as the title theme to the Sega Ages collection of remade Sega Genesis and Master System Sonic titles, and in the opening theme to Sonic Boom: Rise of Lyric, a Japanese anime television series based on Sonic. The Master System port was developed by Sega Technical Institute and published in North America by Acclaim Entertainment as Sonic the Hedgehog (1991) in 1992, and in PAL regions by Ocean Software in 1993 as Sega's Greatest Hits Sonic. A Game Gear port was released by Ocean in 1992. A port for the Sega CD, titled Sega CD's Greatest Sonic Hits (Sega Mega-CD's Sonic Hits in Europe) and published by Sega of Europe, was\\
Nucleus (13b) & A version of Sonic the Hedgehog was developed by Ancient and released in 1991 for Sega's 8 - bit consoles, the Master System and Game Gear. While the game struggled to sell and was poorly received, it represented an experiment in Sonic game capability by requiring that the game's code be 1,000 times faster as a response to Sega's attempts to catch up to Nintendo. This early version shares only a few very minor features with the official Sonic the Hedgehog game released three and a half years later for the Super Nintendo Entertainment System by Sonic Team, notably features from Sonic Spinball (working as a gravity suit for the player and a non-invincible roll ability)are retained. Though it is possible that the data from this very early build was also used in a test build of Sonic's follow-up entries to Sonic Gems: Team Sonic Racing 2, a point that has since been supported by Victor Coelho, one of the creators of the ROM, in the Sonic Mega Collection Plus special discs. Unfortunately, and uniquely compared to its very similarly named paleo relative of 1991 in Mobile Suit Gundam: MSI,both 2D and its computer version are far less distinctly different than the 2D Game Gear version even while being different to and games.

There were also Sonic propositions for a Game Boy Advance port due to which Sega developed a prototype for it using Game Studio, an in-house game-\\
\bottomrule
\end{tabular}
\caption{\label{tab:qua3} Qualitative Example 3.}
\end{table*}

\begin{table*}

\small
    \centering
    \begin{tabular}{lp{13cm}}
\toprule
Prompt &  Writing for The A.V. Club, television reviewer Hayden Childs commented that " Exit Through the Kwik - E - Mart "\\ 
\midrule
CD (1.5b) &   Writing for The A.V. Club, television reviewer Hayden Childs commented that " Exit Through the Kwik - E - Mart " is "the best episode of the show yet." I agree with Childs. This episode is a masterpiece of the kind of television that The Simpsons is so often lauded as having. The story, written by showrunner Mike Scully and starring the always excellent John Di Mambro as the voiceover narrator of "Exit Through The Kwik- E - Mart," has the feel and feel-ness of the classic "Bart to the Future," but with the wit of a modern Simpsons. The story, in its first two acts, follows Bart's attempts to find the perfect gift for a girl he's crushing on, while his parents are away at the hospital for the birth of Bart's sister Lisa (the episode was written before Lisa was conceived, so Bart is not aware that Lisa's father, Homer Simpson, will one day marry Marge Simpson and become Homer Simpson, and thus the father of Lisa and the future Homer). The third and fourth acts are devoted entirely to the birth and early childhood years of Lisa, as Bart is left behind at home to raise her while Homer is off working for the nuclear plant, leaving Bart with the task of looking after her while he goes out and gets drunk and/or high and/or whatever else happens when you have \\

Nucleus (1.5b) &   Writing for The A.V. Club, television reviewer Hayden Childs commented that " Exit Through the Kwik - E - Mart " is one of " the fucking worst shows ever made on television," which makes sense since it made no sense at all. Why get all obsessive about specific comic book characters when you can "relate" with people from any fictional realm? Always good to know. The one-shot from Grant Morrison and Darick Robertson sets up the Missing Link story of the issue, a side narrative that follows a resistance mission into a parallel universe and leads to a spooky-as-hell finale. Next week: The Walking Dead So why would Morrison want to direct a feature film, or at least in such a genre? Well, he's already turned his attention to television and presumably want to turn his hand to the tiny screen again. That's where this comes in—the TV series he's being given the job to write and direct is The Walking Dead, the \$1.2 billion-selling zombie apocalypse franchise on AMC that's probably the most viewed zombie phenomenon on the planet (have you seen the true facts? Those ratings really aren't close) The show is about a kid from the first wave of zombies, a few years after the zombie apocalypse has come and gone. He doesn't know anything about zombies. He just knows that you can't leave the woods around a campground\\

CD (13b) & Writing for The A.V. Club, television reviewer Hayden Childs commented that " Exit Through the Kwik - E - Mart " is "an entertaining, if not particularly insightful documentary about graffiti artist Kaws, whose work has become increasingly commercial and mainstream." He added that "Kawasaki is a likable enough subject, and the film does manage to capture some interesting footage of Kaws at work in his studio and around New York. But the documentary never quite manages to explain what exactly it's trying to say about art or commerce or the nature of fame." He concluded by writing that "Exit Through the Kwik - E - Mart is a decent enough documentary about an intriguing figure. But it's also an example of how, even with a talented subject and some decent access, the best documentaries still can't quite escape their limitations as entertainment. [4/10]" [6]

For The Hollywood Reporternotablet review, critic Matt Zoller Seitz gave the documentary two-and-half-stars out of four and wrote that "Kaws, the graffiti artist and pop-culture phenom who's become the most visible ambassador for street art since Jean Michel-Basquiat and Keith Haring, has a documentary made about him by his friend and business associate, the director and photographer Andrew Droz Palermo. The result, Exit Through the\\

Nucleus (1.5b) & Writing for The A.V. Club, television reviewer Hayden Childs commented that " Exit Through the Kwik - E - Mart " is easily one of the funniest episodes in Rick and Morty history. "They compete for the chance to design a McDonald's restaurant with outlandish designs and terrible burgers," Childs writes, adding "the encounter between Rick and his grandson turns the intergalactic adventure on its head when Morty attempts to sex his planet-derailing Grandpa." To date, this episode has been called one of the Saint-140 standouts of the series.
 In the Season Two episode "Mortynight Run", Rick recommends that Morty to sleep in a Kwik-E-Mart, which he does every year on his birthday.
 In the Season Three episode "Uter short" the Kwik-E-Mart from Kwik-E-Mart is seen under construction.

References

External links
 Official website of the makers of Rick and Morty. (archived)
 Official Website of WDAS TV Store
 [ Kwik-E-Mart Episode Listening Center] (web archive)
\\

\bottomrule
\end{tabular}
\caption{\label{tab:qua4} Qualitative Example 4.}
\end{table*}

\begin{table*}
\small 
    \centering
    \begin{tabular}{lp{13cm}}
\toprule
Prompt &   DreamWorks contracted Amazon.com to create the official website, marking the first time that Amazon had created a special section devoted to a feature film. The website\\ 
\midrule
CD (1.5b) &    DreamWorks contracted Amazon.com to create the official website, marking the first time that Amazon had created a special section devoted to a feature film. The website features the official trailers, cast and credits of all five DreamWorks features. The website is also the first time Amazon has provided the official website of any feature film to the public, as they had previously only released the film's trailers and posters to their employees and select press. The official site is available in both a standard version, which is the one you are currently reading, or an "Enhanced" version that has additional content and is optimized to load quickly and load the most information possible on a mobile screen. The enhanced site also includes a "DreamWorks Home Page," a section that provides a more personalized experience and offers a "DreamWorks Home Screen." The site will be available to all Amazon Prime customers, but is currently available to Prime customers in the US, the United Kingdom and Germany, with the site's availability expanding to the remaining markets later. The official website for The Croods is currently unavailable in Australia. The site is expected to return to Australia later in 2012. The site is currently unavailable in Brazil, China and the Netherlands, while it will be added in these countries in the coming days, but not all features are yet available. The site is also unavailable for users in India. The website for the animated feature Trolls was also launched today. The site features\\

Nucleus (1.5b) &    DreamWorks contracted Amazon.com to create the official website, marking the first time that Amazon had created a special section devoted to a feature film. The website is also available on AOL Instant Messenger f
or voice-enabled users. The company announced it was doing the project in July but didn't say what kind of publicity it had gathered.

Participating film-makers included students at Amazon's "digital artist lab," which has worked with companies such as Dell, Redwood City, Calif.-based TV maker Vizio and Daniel Crown, a Dominican immigrant who made his name in the 1980s as the lead singer of the Demolition Dog.

DreamWorks already had a large, dedicated online fans site for the film, which stars the popular actors Jason Bateman, Owen Wilson, Christian Bale, and Michelle Pfeiffer, as well as more recent comic-book favorites Emily Blunt and Paul Rudd. The home page had a map of Manhattan with other New Yorkers who would be familiar to the film-viewing public.

From various corners of the Internet

The site is not quite as polished as an official website for an old movie. The 1920s-era design isn't very elegant and Microsoft's Internet Explorer plugin doesn't load smoothly. For example, a rudimentary search function is available, but not very sophisticated. Google is also a huge presence, as users can consult news about the project on Google News\\

CD (13b) & DreamWorks contracted Amazon.com to create the official website, marking the first time that Amazon had created a special section devoted to a feature film. The website featured exclusive interviews and footage, a blog by Dreamworks' Jeffrey Katzenberg and Meg LeFauve and a section where visitors could create a personalized character. The film's website received a nomination for Best Website at the 2008 Streamys.

Release

Marketing

DreamWorks began its campaign for Shrek Forever After on May 7 with a teaser website, featuring the tag line "The adventure continues", and a countdown clock to May 17, when the teaser trailer would premiere on ABC's Jimmy Kimmel Live!. The trailer debuted during the show and was also made available online. The official trailer premiered on the same date. The film's first poster was revealed in July, featuring the four maincharacters in silhouette. A second poster, featuring the four characters in full color and the tag line "The adventure never ends" debuted at San Diego Comic-Con. A series of character posters was also revealed. The first of which featured Fiona, Donkey and Puss in Boots in silhouette and was revealed on August 1. A second character poster featuring Shrek, Princess Fiona and Donkey debuted in August, followed by posters featuring Shreks son, Farquad and the villainous Lord Farquaad in September, and the four characters together in October. The first television spot \\ 

Nucleus (13b) & DreamWorks contracted Amazon.com to create the official website, marking the first time that Amazon had created a special section devoted to a feature film. The website previously served as store and promotional hub for all DreamWorks films.

According to ThinkFilm, the website is sort of like a cross between Backstage and IMDbPro. Users can even take a short quiz and get a free preview of the film. Members of the website are also privy to additional content and promotions.

Other sites frequented by the celebs during their promotional tour were Artsy.net, wellcome.org, Mashable and amfam.com.
http://www.ingorepub.com/

Insight–it’s who you know

by Emily Kreps (If Magazine)

Wal-Mart is watching you. Could that even be an exaggeration? Wal-Mart, the world’s largest retailer, is moving quickly to fill the information void, in providing a signal its general awareness of what goes on from
 brick and mortar location to location.

The chain has equipped what’s essentially online GPS monitoring data of nearly every metric found in traditional retail, and is adding capabilities on the fly as it knows what’s measurable–but still not. While Wal-
Mart isn’t tapping into cook’s sales in the chilled salad department, the ability to \\ 

\bottomrule
\end{tabular}
\caption{\label{tab:qua5} Qualitative Example 5.}
\end{table*}

\begin{table*}
    \small 
    \centering
    \begin{tabular}{lp{13cm}}
\toprule
Prompt &    Churchill has been active since at least 1984 as the co - director of the Denver - based American Indian Movement of Colorado, now an autonomous\\ 
\midrule
CD (1.5b) &     Churchill has been active since at least 1984 as the co - director of the Denver - based American Indian Movement of Colorado, now an autonomous entity. He has also been active as the co - of the American Indian Movement of the Rockies, the American Indians for a Free and Healthy Colorado (AIFHFC), the Colorado Native Youth Coalition (CONYEC), and as an advisor for several other Colorado based groups. In the 1980's, Churchill served on the Colorado Indian Tribal Advisory Committee, the Denver Urban Indian Advisory Council (DUIC), the Denver Indian Youth Advisory Committee (DENIYC) and as the co - director for the Denver Indian Education Project (DENIEF), which is the only Native-led, Native-run school reform initiative of any size.

In the 1990s Churchill co - directed and was co-chair for a group that developed the "Denver Native Youth Initiative (DYNIEI)", a Native youth empowerment and education program. The initiative's mission statement i
s "To provide Native students, youth and their parents and guardians, with culturally-relevant education, leadership and community development programs." The initiative is based at Denver's Indian School of Denver
(ISCED), a Native American boarding and day school that serves over 500 Native American students from grades K-8, with an additional 200 Native students in the ISCD's\\

Nucleus (1.5b) &    Churchill has been active since at least 1984 as the co - director of the Denver - based American Indian Movement of Colorado, now an autonomous organization of the American Indian Movement. He is also active in the Colorado Springs IND at the level of the North American's Assembly for National Congress of American Indians (NANIC) President. He will be addressing the 16th National Congress of the American Indian Movement in Denver to come Thursday-Sunday with a community part on Tuesday and Wednesday. To send a postcard: ucpsachaol.com. Also available by e-mail at above address. Come with a great group - the picket line is not usually organized in the city and there are long lines for tickets to see him at meetings, but I can get two lucky letter writers a ticket. :-) And now for the links: Classic, First book Churchill, Lawrence and Jonassen, R.J. 1976 Memoirs of an Indian Killer Manifestos of the Indian Supreme Court 1959 by www.nativeamericanutcc.net Mountain Justice Party Rocky Mountain Front Indian Shooting Party Militia Freedom for the Chesapeakers Fighting the Apache Indian Wars Tylecote Blood Justice League Savoy Caserotti Gulch Citizens for Human Rights Mountain Justice College of Medicine Colorado Springs American Indian Movement Colorado Springs NANICNANIC Conference in Denver "to demonstrate the goodness of the American Indian of Colorado" \\

CD (13b) & Churchill has been active since at least 1984 as the co - director of the Denver - based American Indian Movement of Colorado, now an autonomous - organization with chapters in several states. He was also active as the co - of the Denver based Colorado American Indian Movement of Colorado, now a defunct - organization with no known successor. Churchill is also known to have been the - director for a short period in 1986 of a now defunct Colorado based American Indian Movement of the American West (AIMWAW) with chapters in several western states, and as the co--director in the mid-80s with the late Robert Red Cloud of a defunct Montana based AIW (AIM of MT). Churchill has been involved with several other American Indian activist organizations in various roles, including as a - member and officer for a short time of a defunct Washington State based American Indians in Defense of Our Mother Earth (AIODOME), as the - and officer of an Oregon based American Indians for Survival and Dignity (AISSD) from the mid-80s until its dissolution, as the co - of an  Idaho based American Indians United for Self Determination and Education (AIUSA) chapter from 1985-86 and again from 1989 until the chapter's dissolution, as a - and officer of a now dissolved Washington based American Indians of North America ( \\ 
Nucleus (13b) & Churchill has been active since at least 1984 as the co - director of the Denver - based American Indian Movement of Colorado, now an autonomous entity, and as a prophet-poet of sorts, performing his "Magic Ma
ce" ritual while chanting a Warrior Poem of... More >>>

Archived News

Tips for nights out in the shadow of the Hotel Colorado's ship:

See a performance by the Macabre Heart Show at the Chuckwagon Saloon or the Havana Bar

Have dinner at Lahti's

Spend sometime drifting around an Avalanche adult beverage or beer garden

Take in a performance by a Wallenberg student ensemble

Holiday toy boutique draws the line to who's left out

STOCKTON - When most kids sit down during the holidays to play with their gifts, a small number of children on the edge of poverty will take a breath and wonder if they will get gifts at all.

There's a secret magic about the Radio City Christmas Spectacular

The United Way of San Joaquin County recently released their list of Christmas wish list toys in the area. Many of the toys can be found around the area for children of all ages. You just have to search a bit.

Christmas Vigil scheduled Dec. 24

In tribute to the birth of Christ, Stockton has included a Christmas Vigil in its city code.\\ 
\bottomrule
\end{tabular}
\caption{\label{tab:qua6} Qualitative Example 6.}
\end{table*}

\begin{table*}
    \centering
    \begin{tabular}{lp{13cm}}
\toprule
Prompt & The old Bintulu airport was built in 1955 in the town centre. It once held the Guinness World Record of nearest airport to town. On 19 December\\ 
\midrule
CD (13b) &   The old Bintulu airport was built in 1955 in the town centre. It once held the Guinness World Record of nearest airport to town. On 19 December 2006 the airport was officially renamed Sultan Haji Omar "Ali Sai
fuddien Saarangkara ibni Al-marhum Sultan Badlishah" International Airport after the current Yang Dipertuan Besar of Bintulu, H.Sultan Omar Ali Saifuddin ibni Sultan Hj Omar Sa'adul Khairi Waddien Al-Muhtadee Billa
h. The old Bintulu airport is now used only for VIP and chartered aircraft. The new Sultan Omar Ali Saifuddien International Airport is located about 10 kilometres away from town, in a new area called the Sultan OA
OS Airport Industrial Estate. It was officially opened by the Yang Di-pertua Negri, Tun Pehin Sri Hj Mohammad Bolkiah Mu'izzaddin Waddien Mujib bin Tun Abdul Jalil Mu'adzam Shah, in November 2007.[2]

BINTUAN (pronunciation: [bantuean]
(help·info), officially Sultan O. A. S International Airport (IATA: BNT),[1]) (Malay: Kepulauan Seribu Tiga Pilihan Raya, "Three States' Territory\\

Nucleus (13b) &    The old Bintulu airport was built in 1955 in the town centre. It once held the Guinness World Record of nearest airport to town. On 19 December 1959 the airport was closed and moved to the now defunct Camp San
ggau military base. After the military base was later closed in 1989, the airport was relocated once again to Bedok. This photo was taken in 2007 at the new Bintulu airport.Lai Wan S Ismail, Bintulu, Malaysia.

Comments

Your email address will not be published. Required fields are marked *

Comment

Name *

Email *

Website

The Mid-Autumn Festival moon is currently high in the Solar precession/North Eastern Star.

Chinese people used to perform ceremonies on the luminescent with peanut oil to seek the goddess of love and marriage, moon goddess, they called Goddess of the Harvest. Goddess of marriage was called Moon goddess,
after all, moon is marital goddess.

xxx

New Year Seas too continues, Chinese people are in state of high expectation with connections to all the New Fate/Lantern Clay Ladder. Along the whole period of Glory Fest period, like many Lantern Clay poles, ther
e is an interval of meaningful interlude.

xxx

Another message in the sky in Chinese dyeing. A blessing of seasonal prosperity:

This season is an\\

\bottomrule
\end{tabular}
\caption{\label{tab:qua7} Qualitative Example 7.}
\end{table*}

\section{Variant of CD: Training the Amateur LM}
\label{app:training}
As we mentioned in \cref{ssec:amateur}, an ideal amateur LM should summarize the failure mode of the expert LM, and we have been using a off-the-shelf amateur LM in the main text (e.g., GPT-2 small, OPT-125m). Here, we experiment with learning an amateur model that mimics the  degenerate behavior of the expert LM. Precisely, we first randomly sample some prompt of different length from wikipedia dataset, and generate training data by beam searching the expert LM conditioned on the prompts. This training data is representative of the degeneration in the expert LM, and tends to be highly repetitive. We then prefix-tune \cite{li-liang-2021-prefix} a GPT-2 model on this training data to obtain the final amateur LM. Here, we use prefix-tuning as the lightweight adaptation method which only requires learning and storing a soft prompt of length 10. At decoding time, we just use the prefix-tuned model as the amateur, and apply contrastive decoding in \cref{ssec:contrastive}. We denote this variant of CD as \emph{beamprefix} and report automatic evaluation results in \cref{tab:wikitext}, \cref{tab:wikinews}, and \cref{tab:story}. 

We also include human evaluation results, which compares the beamprefix variant of CD with nucleus sampling results. As shown in \cref{tab_app:human_eval}, we find that CD (beamprefix) also attain significantly better performance than nucleus sampling. 

\begin{table*}[]
\centering
\resizebox{1.0\textwidth}{!}{
\begin{tabular}{lcc|ccc|ccccccllllll}
& & &  \multicolumn{3}{c}{coherence} & \multicolumn{3}{c}{fluency} \\
& CD & Baseline  & CD is better &     same  &    Baseline is better &  CD is better &     same  &    Baseline is better & \\

\toprule

\multirow{4}{*}{\rotatebox{90}{wikitext}} 
& CD (GPT-2 XL) &  nucleus (GPT-2 XL) & 0.714 & 0.083 & 0.202 & 0.548 & 0.083 & 0.369 \\
& CD (beamprefix) &  nucleus (GPT-2 XL) &  0.742 & 0.081 & 0.177 & 0.551 & 0.141 & 0.308 \\ 
\midrule
\multirow{4}{*}{\rotatebox{90}{wikinews}} 
& CD (GPT-2 XL) &  nucleus (GPT-2 XL) & 0.708 & 0.042 & 0.25 & 0.583 & 0.12 & 0.297 \\ 
& CD (beamprefix) & nucleus (GPT-2 XL) & 0.62 & 0.214 & 0.167 & 0.589 & 0.271 & 0.141 \\
\midrule
\multirow{4}{*}{\rotatebox{90}{story}} 
& CD (GPT-2 XL) &  nucleus (GPT-2 XL) & 0.636 & 0.045 & 0.318 & 0.404 & 0.106 &  0.49 \\ 
& CD (beamprefix) & nucleus (GPT-2 XL) & 0.662 & 0.035 & 0.303 & 0.46 & 0.157 & 0.384 \\
\bottomrule

\end{tabular}
}
\caption{\label{tab_app:human_eval} Human evaluation results for wikipedia, wikinews, story  datasets. We describe the details of CD (beamprefix) in \cref{app:training}.}
\end{table*}

\end{document}